\documentclass[graybox]{svmult}

\usepackage{type1cm}        
\usepackage{makeidx}         
\usepackage{graphicx}        
\usepackage{multicol}        
\usepackage[bottom]{footmisc}

\usepackage{newtxtext}       %
\usepackage{newtxmath}       

\usepackage{xspace} 
\usepackage{multirow}
\usepackage{booktabs}

\newcommand{\system}{\textsc{HYPE}\xspace}
\newcommand{\systemdesc}{\textsc{Human eYe Perceptual Evaluation}\xspace}
\newcommand{\systeminf}{$\system_{\infty}$\xspace}
\newcommand{\systemstair}{$\system_{\text{time}}$\xspace}
\newcommand{\taskceleba}{${\text{CelebA-64}}$\xspace}
\newcommand{\taskffhq}{${\text{FFHQ-1024}}$\xspace}
\newcommand{\taskimagenet}{${\text{ImageNet-5}}$\xspace}
\newcommand{\taskcifar}{${\text{CIFAR-10}}$\xspace}
\newcommand{\trunc}{$\text{StyleGAN}_{\text{trunc}}$\xspace}
\newcommand{\notrunc}{$\text{StyleGAN}_{\text{no-trunc}}$\xspace}
\newcommand{\BigGANtrunc}{$\text{BigGan}_{\text{trunc}}$\xspace}
\newcommand{\BigGANnotrunc}{$\text{BigGan}_{\text{no-trunc}}$\xspace}
\newcommand{\website}{https://hype.stanford.edu\xspace}

\newcommand\blfootnote[1]{%
  \begingroup
  \renewcommand\thefootnote{}\footnote{#1}%
  \addtocounter{footnote}{-1}%
  \endgroup
}

\makeindex             

\begin{document}

\title*{Visual Intelligence through Human Interaction}
\author{Ranjay Krishna, Mitchell Gordon, Li Fei-Fei, Michael Bernstein}
\institute{Ranjay Krishna \at Stanford University, \email{ranjaykrishna@cs.stanford.edu}
\and Mitchell Gordon \at Stanford University, \email{mgord@cs.stanford.edu}
\and Li Fei-Fei \at Stanford University, \email{feifeili@cs.stanford.edu}
\and Michael Bernstein \at Stanford University, \email{msb@cs.stanford.edu}}
%
%
\maketitle

\abstract*{

Over the last decade, Computer Vision, the branch of Artificial Intelligence aimed at understanding the visual world, has evolved from simply recognizing objects in images to describing pictures, answering questions about images, aiding robots maneuver around physical spaces and even generating novel visual content. As these tasks and applications have modernized, so too has the reliance on more data, either for model training or for evaluation. In this chapter, we demonstrate that novel interaction strategies can enable new forms of data collection and evaluation for Computer Vision.
First, we present a crowdsourcing interface for speeding up paid data collection by an order of magnitude, feeding the data-hungry nature of modern vision models.
Second, we explore a method to increase volunteer contributions using automated social interventions.
Third, we develop a system to ensure human evaluation of generative vision models are reliable, affordable and grounded in psychophysics theory.
We conclude with future opportunities for Human-Computer Interaction to aid Computer Vision.
\blfootnote{\small{This is a preprint of the following chapter: Ranjay Krishna, Mitchell Gordon, Li Fei-Fei, Michael Bernstein, Visual Intelligence through Human Interaction, published in Artificial Intelligence for Human Computer Interaction: A Modern Approach, edited by Yang Li and Otmar Hilliges, 2021, Springer reproduced with permission of Springer Nature. The final authenticated version is available online at:https://doi.org/10.1007/978-3-030-82681-9}}}

\abstract{

Over the last decade, Computer Vision, the branch of Artificial Intelligence aimed at understanding the visual world, has evolved from simply recognizing objects in images to describing pictures, answering questions about images, aiding robots maneuver around physical spaces and even generating novel visual content. As these tasks and applications have modernized, so too has the reliance on more data, either for model training or for evaluation. In this chapter, we demonstrate that novel interaction strategies can enable new forms of data collection and evaluation for Computer Vision.
First, we present a crowdsourcing interface for speeding up paid data collection by an order of magnitude, feeding the data-hungry nature of modern vision models.
Second, we explore a method to increase volunteer contributions using automated social interventions.
Third, we develop a system to ensure human evaluation of generative vision models are reliable, affordable and grounded in psychophysics theory.
We conclude with future opportunities for Human-Computer Interaction to aid Computer Vision.
\blfootnote{\small{This is a preprint of the following chapter: Ranjay Krishna, Mitchell Gordon, Li Fei-Fei, Michael Bernstein, Visual Intelligence through Human Interaction, published in Artificial Intelligence for Human Computer Interaction: A Modern Approach, edited by Yang Li and Otmar Hilliges, 2021, Springer reproduced with permission of Springer Nature. The final authenticated version is available online at:https://doi.org/10.1007/978-3-030-82681-9}}}

\section{Introduction}
\label{sec:1}
Today, Computer Vision applications are ubiquitous. They filter our pictures, control our car, aid medical experts in disease diagnosis, analyze sports games, and even generate complete new content. This recent emergence of Computer Vision tools has been made possible because of a shift in the underlying techniques used to train models; this shift has transferred attention away from hand engineered features~\cite{lowe1999object,dalal2005histograms} towards deep learning~\cite{deng2009imagenet,krizhevsky2012imagenet}. With deep learning techniques, vision models have surpassed human performance on fundamental tasks, such as object recognition~\cite{russakovsky2014imagenet}. Today, vision models are capable of an entirely new host of applications, such as generating photo-realistic images~\cite{karras2019style} and 3D spaces~\cite{mildenhall2020nerf}. These tasks have made possible numerous vision-powered applications~\cite{zhang2020vid2player,yue2017scenectrl,xia2020crosscast,laielli2019labelar,huang2019sketchforme}.

This shift is also reflective of yet another change in Computer Vision: algorithms are becoming more generic and data has become the primary hurdle in performance. Today's vision models are data-hungry; they feed on large amounts of annotated training data.
In some cases, data needs to be continuously annotated in order to evaluate models; for new tasks such as image generation, model-generated images can only be evaluated if people provide realism judgments. 
To support data needs, Computer Vision has relied on a specific pipeline for data collection --- one that focuses on manual labeling using online crowdsourcing platforms such as Amazon Mechanical Turk~\cite{deng2009imagenet,visualgenome}.

Unfortunately, this data collection pipeline has numerous limiting factors. First, crowdsourcing can be insufficiently scalable, and it remains too expensive for use in the production of many industry-size datasets~\cite{josephy2013crowdscale}. Cost is bound to the amount of work completed per minute of effort, and existing techniques for speeding up labeling are not scaling as quickly as the volume of data we are now producing that must be labeled~\cite{thomee2016yfcc100m}.
Second, while cost issues may be mitigated by relying on volunteer contributions, it remains unclear how best to incentivize such contributions. Even though there has been a lot of work in Social Psychology exploring strategies to incentivize volunteer contributions to online communities~\cite{kraut2011encouraging,burkemembership,markey2000bystander,darley1968bystander,Yang:2017:PTG:3171581.3134749,info:doi/10.2196/jmir.3558}, it remains unclear how we can employ such strategies to develop automated mechanisms that incentivize volunteer data annotation useful for Computer Vision.
Third, existing data annotation methods are ad-hoc, each executed in idiosyncrasy without proof of reliability or grounding to theory, resulting in high variance in their estimates~\cite{salimans2016improved, denton2015deep, olsson2018skill}. While high-variance in labels might be tolerable when collecting training data, it becomes debilitating when such ad-hoc methods are used to evaluate models.

Human-Computer Interaction's opportunity is to look to novel interaction strategies to break this away from traditional data collection pipeline.
In this chapter, we showcase three projects~\cite{krishna2016embracing,park2019ai,zhou2019hype}, that have helped modern Computer Vision data needs. The first two projects introduce new training data collection interfaces~\cite{krishna2016embracing} and interactions~\cite{park2019ai} while the third introduces a reliable system for evaluating vision models with humans~\cite{zhou2019hype}. Our contributions (1) speed up data collection by an order of magnitude in terms of speed and cost, (2) incentivize volunteer contributions to provide labels through conversational interactions over social media, and (3) capacitate reliable human evaluation of vision models.

In the first section, we highlight work that accelerates human interactions in microtask crowdsourcing, a core process through which computer vision and machine learning datasets are predominantly curated~\cite{krishna2016embracing}.
Microtask crowdsourcing has enabled dataset advances in social science and machine learning, but existing crowdsourcing schemes are too expensive to scale up with the expanding volume of data. To scale and widen the applicability of crowdsourcing, we present a technique that produces extremely rapid judgments for binary and categorical labels. Rather than punishing all errors, which causes workers to proceed slowly and deliberately, our technique speeds up workers' judgments to the point where errors are acceptable and even expected. We demonstrate that it is possible to rectify these errors by randomizing task order and modeling response latency. We evaluate our technique on a breadth of common labeling tasks such as image verification, word similarity, sentiment analysis and topic classification. Where prior work typically achieves a 0.25$\times$ to 1$\times$ speedup over fixed majority vote, our approach often achieves an order of magnitude (10$\times$) speedup.

In the second section, we turn our attention from paid crowdsourcing to volunteer contributions; we explore how to design social interventions to improve volunteer contributions when curating datasets~\cite{park2019ai}.
To support the massive data requirements of modern supervised machine learning algorithms, crowdsourcing systems match volunteer contributors to appropriate tasks. Such systems learn \textit{what} types of tasks contributors are interested to complete. In this paper, instead of focusing on \textit{what} to ask, we focus on learning \textit{how} to ask: how to make relevant and interesting requests to encourage crowdsourcing participation. 
We introduce a new technique that augments questions with learning-based request strategies drawn from social psychology. 
We also introduce a contextual bandit algorithm to select which strategy to apply for a given task and contributor.
We deploy our approach to collect volunteer data from Instagram for the task of visual question answering, an important task in computer vision and natural language processing that has enabled numerous human-computer interaction applications.
For example, when encountering a user's Instagram post that contains the ornate Trevi Fountain in Rome, our approach learns to augment its original raw question ``Where is this place?'' with image-relevant compliments such as ``What a great statue!'' or with travel-relevant justifications such as ``I would like to visit this place'', increasing the user's likelihood of answering the question and thus providing a label. We deploy our agent on Instagram to ask questions about social media images, finding that the response rate improves from $15.8\%$ with unaugmented questions to $30.54\%$ with baseline rule-based strategies and to $58.1\%$ with learning-based strategies.

Finally, in the third section, we spotlight our work on constructing a reliable human evaluation system for generative computer vision models~\cite{zhou2019hype}.
Generative models often use human evaluations to measure the perceived quality of their outputs. Automated metrics are noisy indirect proxies, because they rely on heuristics or pretrained embeddings.
However, up until now, direct human evaluation strategies have been ad-hoc, neither standardized nor validated.
Our work establishes a gold standard human benchmark for generative realism.
We construct \systemdesc (\system), a human benchmark that is \textit{grounded} in psychophysics research in perception, \textit{reliable} across different sets of randomly sampled outputs from a model, able to produce~\textit{separable} model performances, and \textit{efficient} in cost and time. We introduce two variants: one that measures visual perception under adaptive time constraints to determine the threshold at which a model's outputs appear real (e.g. $250$ms), and the other a less expensive variant that measures human error rate on fake and real images sans time constraints.
We test \system across six state-of-the-art generative adversarial networks and two sampling techniques on conditional and unconditional image generation using four datasets: CelebA, FFHQ, CIFAR-10, and ImageNet. 
We find that \system can track the relative improvements between models, and we confirm via bootstrap sampling that these measurements are consistent and replicable.

\section{Data annotation by speeding up human interactions}
\label{sec:2}

Social science~\cite{kittur2008crowdsourcing, mason2012conducting}, interactive systems~\cite{fast2014emergent, kumar2013webzeitgeist} and machine learning~\cite{deng2009imagenet, lin2014microsoft} are becoming more and more reliant on large-scale, human-annotated data. Increasingly large annotated datasets have unlocked a string of social scientific insights~\cite{gilbert2009predicting, burke2013using} and machine learning performance improvements~\cite{krizhevsky2012imagenet, girshick2014rich, vinyals2014show}. One of the main enablers of this growth has been microtask crowdsourcing~\cite{snow2008cheap}. Microtask crowdsourcing marketplaces such as Amazon Mechanical Turk offer a scale and cost that makes such annotation feasible. As a result, companies are now using crowd work to complete hundreds of thousands of tasks per day~\cite{marcuswaran}.

However, even microtask crowdsourcing can be insufficiently scalable, and it remains too expensive for use in the production of many industry-size datasets~\cite{josephy2013crowdscale}. Cost is bound to the amount of work completed per minute of effort, and existing techniques for speeding up labeling (reducing the amount of required effort) are not scaling as quickly as the volume of data we are now producing that must be labeled~\cite{thomee2016yfcc100m}. To expand the applicability of crowdsourcing,
the number of items annotated per minute of effort needs to increase substantially.

\begin{figure}[h]
\centering
\includegraphics[width=\linewidth]{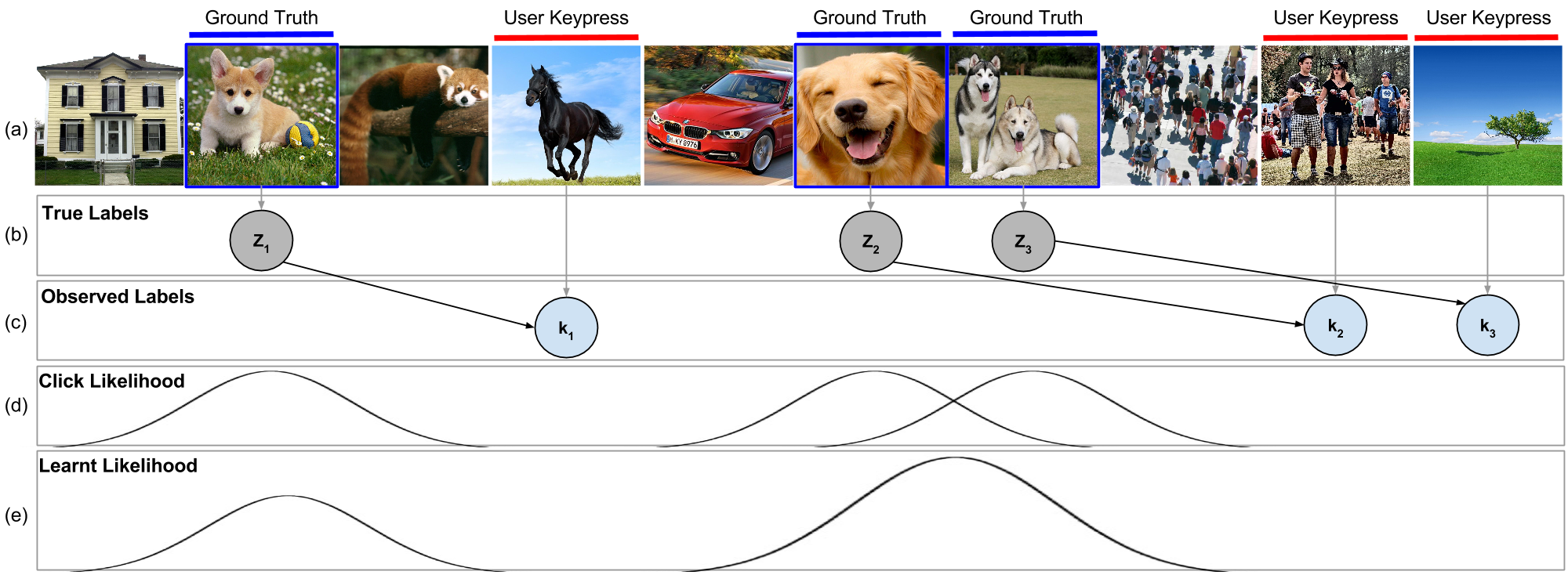}
\caption{(a) Images are shown to workers at 100ms per image. Workers react whenever they see a dog. (b) The true labels are the ground truth dog images. (c) The workers' keypresses are slow and occur several images after the dog images have already passed. We record these keypresses as the observed labels. (d) Our technique models each keypress as a delayed Gaussian to predict (e) the probability of an image containing a dog from these observed labels.}
\label{fig:pull_tsvp}
\end{figure}

In this paper, we focus on one of the most common classes of crowdsourcing tasks \cite{ipeirotis2010analyzing}: binary annotation. These tasks are yes-or-no questions, typically identifying whether or not an input has a specific characteristic. Examples of these types of tasks are topic categorization (e.g., ``Is this article about finance?'')~\cite{schapire2000boostexter}, image classification (e.g., ``Is this a dog?'')~\cite{deng2009imagenet, lin2014microsoft, li2003detecting}, audio styles~\cite{seetharaman2014crowdsourcing} and emotion detection~\cite{li2003detecting} in songs (e.g., ``Is the music calm and soothing?''), word similarity (e.g., ``Are \textit{shipment} and \textit{cargo} synonyms?'')~\cite{miller1991contextual} and sentiment analysis (e.g.,\ ``Is this tweet positive?'')~\cite{pang2008opinion}. 

Previous methods have sped up binary classification tasks by minimizing worker error.
A central assumption behind this prior work has been that workers make errors because they are not trying hard enough (e.g.,\ ``a lack of expertise, dedication [or] interest''~\cite{sheng2008get}).
Platforms thus punish errors harshly, for example by denying payment. Current methods calculate the minimum redundancy necessary to be confident that errors have been removed~\cite{sheng2008get, smyth1994knowledge, smyth1995inferring}. These methods typically result in a 0.25$\times$ to 1$\times$ speedup beyond a fixed majority vote~\cite{peng2010decision,russakovsky2015best,sheng2008get,karger2014budget}.

We take the opposite position: that designing the task to encourage some error, or even make errors inevitable, can produce far greater speedups.
Because platforms strongly punish errors, workers carefully examine even straightforward tasks to make sure they do not represent edge cases~\cite{martin2014being, irani2013turkopticon}. The result is slow, deliberate work. 
We suggest that there are cases where we can encourage workers to move quickly by telling them that making some errors is acceptable.
Though individual worker accuracy decreases, we can recover from these mistakes post-hoc algorithmically (Figure~\ref{fig:pull_tsvp}).

We manifest this idea via a crowdsourcing technique in which workers label a rapidly advancing stream of inputs. Workers are given a binary question to answer, and they observe as the stream automatically advances via a method inspired by rapid serial visual presentation (RSVP)~\cite{li2002rapid, fei2007we}. Workers press a key whenever the answer is ``yes'' for one of the stream items. Because the stream is advancing rapidly, workers miss some items and have delayed responses. However, workers are reassured that the requester expects them to miss a few items. To recover the correct answers, the technique randomizes the item order for each worker and model workers' delays as a normal distribution whose variance depends on the stream's speed. For example, when labeling whether images have a ``barking dog'' in them, a self-paced worker on this task takes 1.7s per image on average. With our technique, workers are shown a stream at 100ms per image. The technique models the delays experienced at different input speeds and estimates the probability of intended labels from the key presses.

We evaluate our technique by comparing the total worker time necessary to achieve the same precision on an image labeling task as a standard setup with majority vote.
The standard approach takes three workers an average of 1.7s each for a total of 5.1s.
Our technique achieves identical precision (97\%) with five workers at 100ms each, for a total of 500ms of work. The result is an order of magnitude speedup of 10$\times$.

This relative improvement is robust across both simple tasks, such as identifying dogs, and complicated tasks, such as identifying ``a person riding a motorcycle'' (interactions between two objects) or ``people eating breakfast'' (understanding relationships among many objects). We generalize our technique to other tasks such as word similarity detection, topic classification and sentiment analysis. Additionally, we extend our method to categorical classification tasks through a ranked cascade of binary classifications. Finally, we test workers' subjective mental workload and find no measurable increase.

Overall, we make the following contributions: (1) We introduce a rapid crowdsourcing technique that makes errors normal and even inevitable. We show that it can be used to effectively label large datasets by achieving a speedup of an order of magnitude on several binary labeling crowdsourcing tasks. (2) We demonstrate that our technique can be generalized to multi-label categorical labeling tasks, combined independently with existing optimization techniques, and deployed without increasing worker mental workload.

\subsection{Related Work}
The main motivation behind our work is to provide an environment where humans can make decisions quickly. We encourage a margin of human error in the interface that is then rectified by inferring the true labels algorithmically. In this section, we review prior work on crowdsourcing optimization and other methods for motivating contributions. Much of this work relies on artificial intelligence techniques: we complement this literature by changing the crowdsourcing interface rather than focusing on the underlying statistical model.

Our technique is inspired by rapid serial visual presentation (RSVP), a technique for consuming media rapidly by aligning it within the foveal region and advancing between items quickly~\cite{li2002rapid, fei2007we}. RSVP has already been proven to be effective at speeding up reading rates~\cite{wobbrock2002webthumb}. RSVP users can react to a single target image in a sequence of images even at 125ms per image with 75\% accuracy~\cite{potter1976short}. However, when trying to recognize concepts in images, RSVP only achieves an accuracy of 10\% at the same speed~\cite{potter1969recognition}. In our work, we integrate multiple workers' errors to successfully extract true labels.

Many previous papers have explored ways of modeling workers to remove bias or errors from ground truth labels~\cite{whitehill2009whose, welinder2010multidimensional, zhou2012learning, peng2010decision, ipeirotis2010quality}. For example, an unsupervised method for judging worker quality can be used as a prior to remove bias on binary verification labels~\cite{ipeirotis2010quality}. Individual workers can also be modeled as projections into an open space representing their skills in labeling a particular image~\cite{whitehill2009whose}. Workers may have unknown expertise that may in some cases prove adversarial to the task. Such adversarial workers can be detected by jointly learning the difficulty of labeling a particular datum along with the expertises of workers~\cite{welinder2010multidimensional}. Finally, a generative model can be used to model workers' skills by minimizing the entropy of the distribution over their labels and the unknown true labels~\cite{zhou2012learning}. We draw inspiration from this literature, calibrating our model using a similar generative approach to understand worker reaction times. We model each worker's reaction as a delayed Gaussian distribution.

In an effort to reduce cost, many previous papers have studied the tradeoffs between speed (cost) and accuracy on a wide range of tasks~\cite{wah2014similarity, branson2010visual, wah2011multiclass, russakovsky2014imagenet}. Some methods estimate human time with annotation accuracy to jointly model the errors in the annotation process~\cite{wah2014similarity, branson2010visual, wah2011multiclass}. Other methods vary both the labeling cost and annotation accuracy to calculate a tradeoff between the two~\cite{jain2013predicting, deng2014scalable}. Similarly, some crowdsourcing systems optimize a budget to measure confidence in worker annotations~\cite{karger2011budget, karger2014budget}. Models can also predict the redundancy of non-expert labels needed to match expert-level annotations~\cite{sheng2008get}. Just like these methods, we show that non-experts can use our technique and provide expert-quality annotations; we also compare our methods to the conventional majority-voting annotation scheme.

Another perspective on rapid crowdsourcing is to return results in real time, often by using a retainer model to recall workers quickly~\cite{bernstein2011crowds, lasecki2011real,laput2015zensors}. Like our approach, real-time crowdsourcing can use algorithmic solutions to combine multiple in-progress contributions~\cite{lasecki2012real}. These systems' techniques could be fused with ours to create crowds that can react to bursty requests. 

\begin{figure}[h]
\centering
\includegraphics[width=\linewidth]{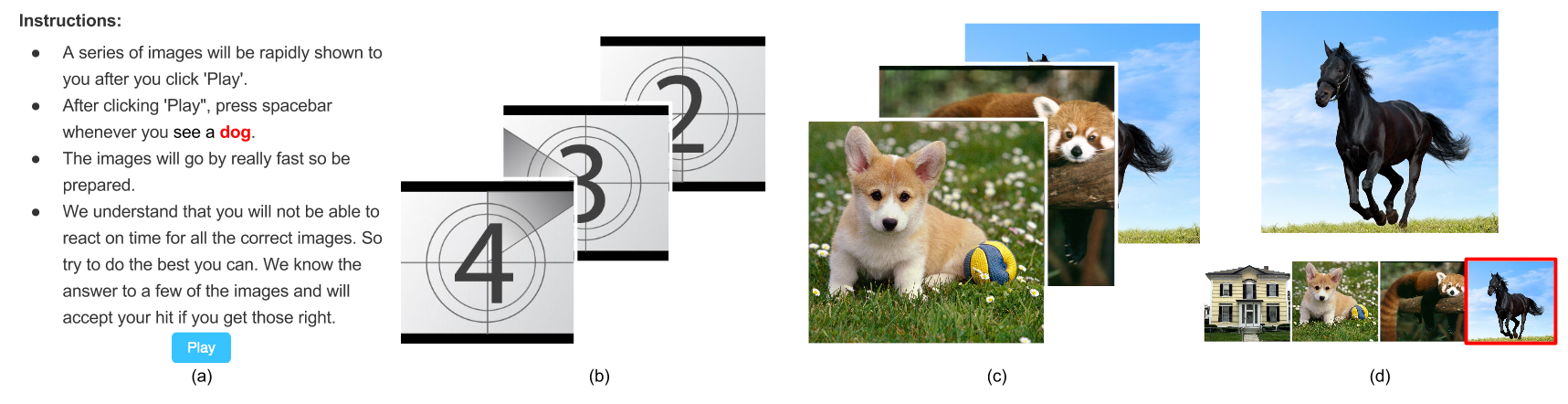}
\caption{(a) Task instructions inform workers that we expect them to make mistakes since the items will be displayed rapidly. (b) A string of countdown images prepares them for the rate at which items will be displayed. (c) An example image of a ``dog'' shown in the stream---the two images appearing behind it are included for clarity but are not displayed to workers. (d) When the worker presses a key, we show the last four images below the stream of images to indicate which images might have just been labeled.}
\label{fig:interface}
\end{figure}

One common method for optimizing crowdsourcing is active learning, which involves learning algorithms that interactively query the user. Examples include training image recognition~\cite{song2011contextualizing} and attribution recognition~\cite{parkash2012attributes} with fewer examples.
Comparative models for ranking attribute models have also optimized crowdsourcing using active learning~\cite{liang2014beyond}. Similar techniques have explored optimization of the ``crowd kernel'' by adaptively choosing the next questions asked of the crowd in order to build a similarity matrix between a given set of data points~\cite{tamuz2011adaptively}.
Active learning needs to decide on a new task after each new piece of data is gathered from the crowd. Such models tend to be quite expensive to compute. Other methods have been proposed to decide on a set of tasks instead of just one task~\cite{vijayanarasimhan2010far}. We draw on this literature: in our technique, after all the images have been seen by at least one worker, we use active learning to decide the next set of tasks. We determine which images to discard and which images to group together and send this set to another worker to gather more information.

Finally, there is a group of techniques that attempt to optimize label collection by reducing the number of questions that must be answered by the crowd. For example, a hierarchy in label distribution can reduce the annotation search space~\cite{deng2014scalable}, and information gain can reduce the number of labels necessary to build large taxonomies using a crowd~\cite{chilton2013cascade, bragg2013crowdsourcing}. Methods have also been proposed to maximize accuracy of object localization in images~\cite{su2012crowdsourcing} and videos~\cite{vondrick2013efficiently}. Previous labels can also be used as a prior to optimize acquisition of new types of annotations~\cite{branson2014active}. One of the benefits of our technique is that it can be used independently of these others to jointly improve crowdsourcing schemes. We demonstrate the gains of such a combination in our evaluation.

\subsection{Error-Embracing Crowdsourcing}
\label{sec:interface_design}
Current microtask crowdsourcing platforms like Amazon Mechanical Turk incentivize workers to avoid rejections~\cite{irani2013turkopticon, martin2014being}, resulting in slow and meticulous work. But is such careful work necessary to build an accurate dataset? In this section, we detail our technique for rapid crowdsourcing by encouraging less accurate work.

The design space of such techniques must consider which tradeoffs are acceptable to make. The first relevant dimension is accuracy. When labeling a large dataset (e.g., building a dataset of ten thousand articles about housing), \textit{precision} is often the highest priority: articles labeled as on-topic by the system must in fact be about housing. \textit{Recall}, on the other hand, is often less important, because there is typically a large amount of available unlabeled data: even if the system misses some on-topic articles, the system can label more items until it reaches the desired dataset size. We thus develop an approach for producing high precision at high speed, sacrificing some recall if necessary.

The second design dimension involves the task characteristics. Many large-scale crowdsourcing tasks involve closed-ended responses such as binary or categorical classifications. These tasks have two useful properties. First, they are time-bound by users' perception and cognition speed rather than motor (e.g., pointing, typing) speed~\cite{cheng2015measuring}, since acting requires only a single button press. Second, it is possible to aggregate responses automatically, for example with majority vote. Open-ended crowdsourcing tasks such as writing~\cite{bernstein2010soylent} or transcription are often time-bound by data entry motor speeds and cannot be automatically aggregated. Thus, with our technique, we focus on closed-ended tasks.

\subsubruninhead{Rapid crowdsourcing of binary decision tasks}
Binary questions are one of the most common classes of crowdsourcing tasks. Each yes-or-no question gathers a label on whether each item has a certain characteristic. In our technique, rather than letting workers focus on each item too carefully, we display each item for a specific period of time before moving on to the next one in a rapid slideshow. For example, in the context of an image verification task, we show workers a stream of images and ask them to press the spacebar whenever they see a specific class of image. In the example in Figure~\ref{fig:interface}, we ask them to react whenever they see a ``dog.''

The main parameter in this approach is the length of time each item is visible. To determine the best option, we begin by allowing workers to work at their own pace. This establishes an initial average time period, which we then slowly decrease in successive versions until workers start making mistakes~\cite{cheng2015measuring}. Once we have identified this error point, we can algorithmically model workers' latency and errors to extract the true labels.

To avoid stressing out workers, it is important that the task instructions convey the nature of the rapid task and the fact that we expect them to make some errors.  Workers are first shown a set of instructions (Figure~\ref{fig:interface}(a)) for the task. They are warned that reacting to every single correct image on time is not feasible and thus not expected. We also warn them that we have placed a small number of items in the set that we know to be positive items. These help us calibrate each worker's speed and also provide us with a mechanism to reject workers who do not react to any of the items.

Once workers start the stream (Figure~\ref{fig:interface}(b)), it is important to prepare them for pace of the task. We thus show a film-style countdown for the first few seconds that decrements to zero at the same interval as the main task. Without these countdown images, workers use up the first few seconds getting used to the pace and speed. Figure~\ref{fig:interface}(c) shows an example ``dog'' image that is displayed in front of the user. The dimensions of all items (images) shown are held constant to avoid having to adjust to larger or smaller visual ranges.

When items are displayed for less than 400ms, workers tend to react to all positive items with a delay. If the interface only reacts with a simple confirmation when workers press the spacebar, many workers worry that they are too late because another item is already on the screen. Our solution is to also briefly display the last four items previously shown when the spacebar is pressed, so that workers see the one they intended and also gather an intuition for how far back the model looks. For example, in Figure~\ref{fig:interface}(d), we show a worker pressing the spacebar on an image of a horse. We anticipate that the worker was probably delayed, and we display the last four items to acknowledge that we have recorded the keypress. We ask all workers to first complete a qualification task in which they receive feedback on how quickly we expect them to react. They pass the qualification task only if they achieve a recall of 0.6 and precision of 0.9 on a stream of 200 items with 25 positives. We measure precision as the fraction of worker reactions that were within 500ms of a positive cue.

\begin{figure}[h]
\centering
\includegraphics[width=\linewidth]{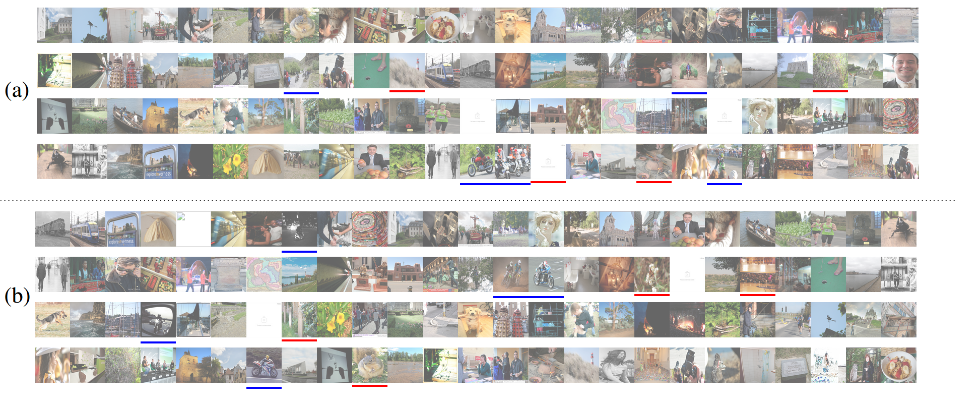}
\caption{Example raw worker outputs from our interface. Each image was displayed for 100ms and workers were asked to react whenever they saw images of ``a person riding a motorcycle.'' Images are shown in the same order they appeared in for the worker. Positive images are shown with a blue bar below them and users' keypresses are shown as red bars below the image to which they reacted.}
\label{fig:qualitative_clicks}
\end{figure}

In Figure~\ref{fig:qualitative_clicks}, we show two sample outputs from our interface. Workers were shown images for 100ms each. They were asked to press the spacebar whenever they saw an image of ``a person riding a motorcycle.'' The images with blue bars underneath them are ground truth images of ``a person riding a motorcycle.'' The images with red bars show where workers reacted. The important element is that red labels are often delayed behind blue ground truth and occasionally missed entirely. Both Figures~\ref{fig:qualitative_clicks}(a) and~\ref{fig:qualitative_clicks}(b) have 100 images each with 5 correct images.

Because of workers' reaction delay, the data from one worker has considerable uncertainty. We thus show the same set of items to multiple workers in different random orders and collect independent sets of keypresses. This randomization will produce a cleaner signal in aggregate and later allow us to estimate the images to which each worker intended to react.

Given the speed of the images, workers are not able to detect every single positive image. For example, the last positive image in Figure~\ref{fig:qualitative_clicks}(a) and the first positive image in Figure~\ref{fig:qualitative_clicks}(b) are not detected. Previous work on RSVP found a phenomenon called ``attention blink''~\cite{broadbent1987detection}, in which a worker is momentarily blind to successive positive images. However, we find that even if two images of ``a person riding a motorcycle'' occur consecutively, workers are able to detect both and react twice (Figures~\ref{fig:qualitative_clicks}(a) and~\ref{fig:qualitative_clicks}(b)). 
If workers are forced to react in intervals of less than 400ms, though, the signal we extract is too noisy for our model to estimate the positive items.

\subsubruninhead{Multi-Class Classification for Categorical Data}
So far, we have described how rapid crowdsourcing can be used for binary verification tasks. Now we extend it to handle multi-class classification. Theoretically, all multi-class classification can be broken down into a series of binary verifications. For example, if there are $N$ classes, we can ask $N$ binary questions of whether an item is in each class. Given a list of items, we use our technique to classify them one class at a time. After every iteration, we remove all the positively classified items for a particular class. We use the rest of the items to detect the next class.

Assuming all the classes contain an equal number of items, the order in which we detect classes should not matter. A simple \textit{baseline approach} would choose a class at random and attempt to detect all items for that class first. However, if the distribution of items is not equal among classes, this method would be inefficient. Consider the case where we are trying to classify items into 10 classes, and one class has 1000 items while all other classes have 10 items. In the worst case, if we classify the class with 1000 examples last, those 1000 images would go through our interface 10 times (once for every class). Instead, if we had detected the large class first, we would be able to classify those 1000 images and they would only go through our interface once. With this intuition, we propose a \textit{class-optimized approach} that classifies the most common class of items first. We maximize the number of items we classify at every iteration, reducing the total number of binary verifications required.

\subsection{Model}
To translate workers' delayed and potentially erroneous actions into identifications of the positive items, we need to model their behavior. We do this by calculating the probability that a particular item is in the positive class given that the user reacted a given period after the item was displayed. By combining these probabilities across several workers with different random orders of the same images, these probabilities sum up to identify the correct items.

We use maximum likelihood estimation to predict the probability of an item being a positive example. Given a set of items $\mathcal{I} = \{I_1, \ldots, I_n\}$, we send them to $W$ workers in a different random order for each. From each worker $w$, we collect a set of keypresses $\mathcal{C}^{w} = \{c_1^w, \ldots, c_k^w\}$ where $w \in W$ and $k$ is the total number of keypresses from $w$. Our aim is to calculate the probability of a given item $P(I_i)$ being a positive example. Given that we collect keypresses from $W$ workers:
\setlength{\belowdisplayskip}{0pt} \setlength{\belowdisplayshortskip}{0pt}
\setlength{\abovedisplayskip}{0pt} \setlength{\abovedisplayshortskip}{0pt}
\vspace{5pt}
\begin{equation}
P(I_i) = \sum_{w} P(I_i | \mathcal{C}^{w}) P(\mathcal{C}^{w})
\end{equation}

where $P(\mathcal{C}) = \prod_{k} P(\mathcal{C}_k)$ is the probability of a particular set of items being keypresses. We set $P(C_k)$ to be constant, asssuming that it is equally likely that a worker might react to any item. Using Bayes' rule:
\vspace{0pt}
\begin{equation}
P(I_i | \mathcal{C}^{w}) = \frac{P(\mathcal{C}^{w} | I_i) P(I_i)}{P(\mathcal{C}^{w})}.
\end{equation}

$P(I_i)$ models our estimate of item $I_i$ being positive. It can be a constant, or it can be an estimate from a domain-specific machine learning algorithm~\cite{kamar2012combining}. For example, to calculate $P(I_i)$, if we were trying to scale up a dataset of ``dog'' images, we would use a small set of known ``dog'' images to train a binary classifier and use that to calculate $P(I_i)$ for all the unknown images. With image tasks, we use a pretrained convolutional neural network to extract image features~\cite{Simonyan14c} and train a linear support vector machine to calculate $P(I_i)$.

We model $P(\mathcal{C}^{w} | I_i)$ as a set of independent keypresses:

\begin{equation}
P(\mathcal{C}^{w} | I_i) = P(c_1^w, \ldots, c_k^w | I_i) = \prod_{k} P(\mathcal{C}^{w}_k | I_i).
\end{equation}

Finally, we model each keypress as a Gaussian distribution $\mathcal{N}(\mu, \sigma)$ given a positive item. We train the mean $\mu$ and variance $\sigma$ by running rapid crowdsourcing on a small set of items for which we already know the positive items. Here, the mean and variance of the distribution are modeled to estimate the delays that a worker makes when reacting to a positive item. 

Intuitively, the model works by treating each keypress as creating a Gaussian ``footprint'' of positive probability on the images about 400ms before the keypress (Figure~\ref{fig:pull_tsvp}). The model combines these probabilities across several workers to identify the images with the highest overall probability.

Now that we have a set of probabilities for each item, we need to decide which ones should be classified as positive. We order the set of items $\mathcal{I}$ according to likelihood of being in the positive class $P(I_i)$. We then set all items above a certain threshold as positive. This threshold is a hyperparameter that can be tuned to trade off precision vs.\ recall. 

In total, this model has two hyperparameters: (1) the threshold above which we classify images as positive and (2) the speed at which items are displayed to the user. We model both hyperparameters in a per-task (image verification, sentiment analysis, etc.) basis. For a new task, we first estimate how long it takes to label each item in the conventional setting with a small set of items. Next, we continuously reduce the time each item is displayed until we reach a point where the model is unable to achieve the same precision as the untimed case.


\begin{figure}[h]
\centering
\sidecaption
\includegraphics[width=0.6\linewidth]{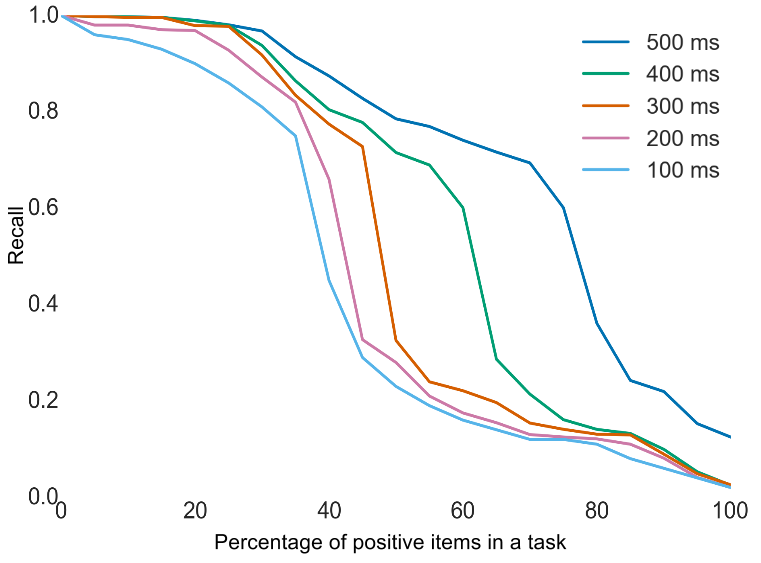}
\caption{We plot the change in recall as we vary percentage of positive items in a task. We experiment at varying display speeds ranging from 100ms to 500ms. We find that recall is inversely proportional to the rate of positive stimuli and not to the percentage of positive items.}
\label{fig:recall_percentage}
\end{figure}

\begin{table}[h]
\centering
\begin{tabular}{l l r r r r r r r}
  \textbf{Task} & & \multicolumn{3}{c}{\textbf{Conventional Approach}} & \multicolumn{3}{c}{\textbf{Our Technique}} & \textbf{Speedup} \\
  \cmidrule(r){3-5}
  \cmidrule(r){6-8}
  & 
  &
  \multicolumn{1}{c}{\textit{Time (s)}} & 
  \multicolumn{1}{c}{\textit{Precision}} & 
  \multicolumn{1}{c}{\textit{Recall}} & 
  \multicolumn{1}{c}{\textit{Time (s)}} & 
  \multicolumn{1}{c}{\textit{Precision}} & 
  \multicolumn{1}{c}{\textit{Recall}} & 
  \multicolumn{1}{c}{}\\
  \hline
 \multirow{4}{*}{Image Verification} & Easy & 1.50 & 0.99 & 0.99 & 0.10 & 0.99 & 0.94 &  \textbf{9.00}$\times$ \\
                                   & Medium & 1.70 & 0.97 & 0.99 & 0.10 & 0.98 & 0.83 & \textbf{10.20}$\times$ \\
                                     & Hard & 1.90 & 0.93 & 0.89 & 0.10 & 0.90 & 0.74 & \textbf{11.40}$\times$ \\
                                     \cmidrule(r){2-9}
                                      & All Concepts & 1.70 & 0.97 & 0.96 & 0.10 & 0.97 & 0.81 & \textbf{10.20}$\times$ \\
 \hline
 Sentiment Analysis & &  4.25 & 0.93 & 0.97 & 0.25 & 0.94 & 0.84 & \textbf{10.20}$\times$ \\
 Word Similarity    & &  6.23 & 0.89 & 0.94 & 0.60 & 0.88 & 0.86 &  \textbf{6.23}$\times$ \\
 Topic Detection    & & 14.33 & 0.96 & 0.94 & 2.00 & 0.95 & 0.81 & \textbf{10.75}$\times$ \\
 \hline
\end{tabular}
\caption{We compare the conventional approach for binary verification tasks (image verification, sentiment analysis, word similarity and topic detection) with our technique and compute precision and recall scores. Precision scores, recall scores and speedups are calculated using 3 workers in the conventional setting. Image verification, sentiment analysis and word similarity used 5 workers using our technique, while topic detection used only 2 workers. We also show the time taken (in seconds) for 1 worker to do each task.}
\label{tab:precision_recall_speedup}
\end{table}

\begin{figure}[h]
\centering
\includegraphics[width=\linewidth]{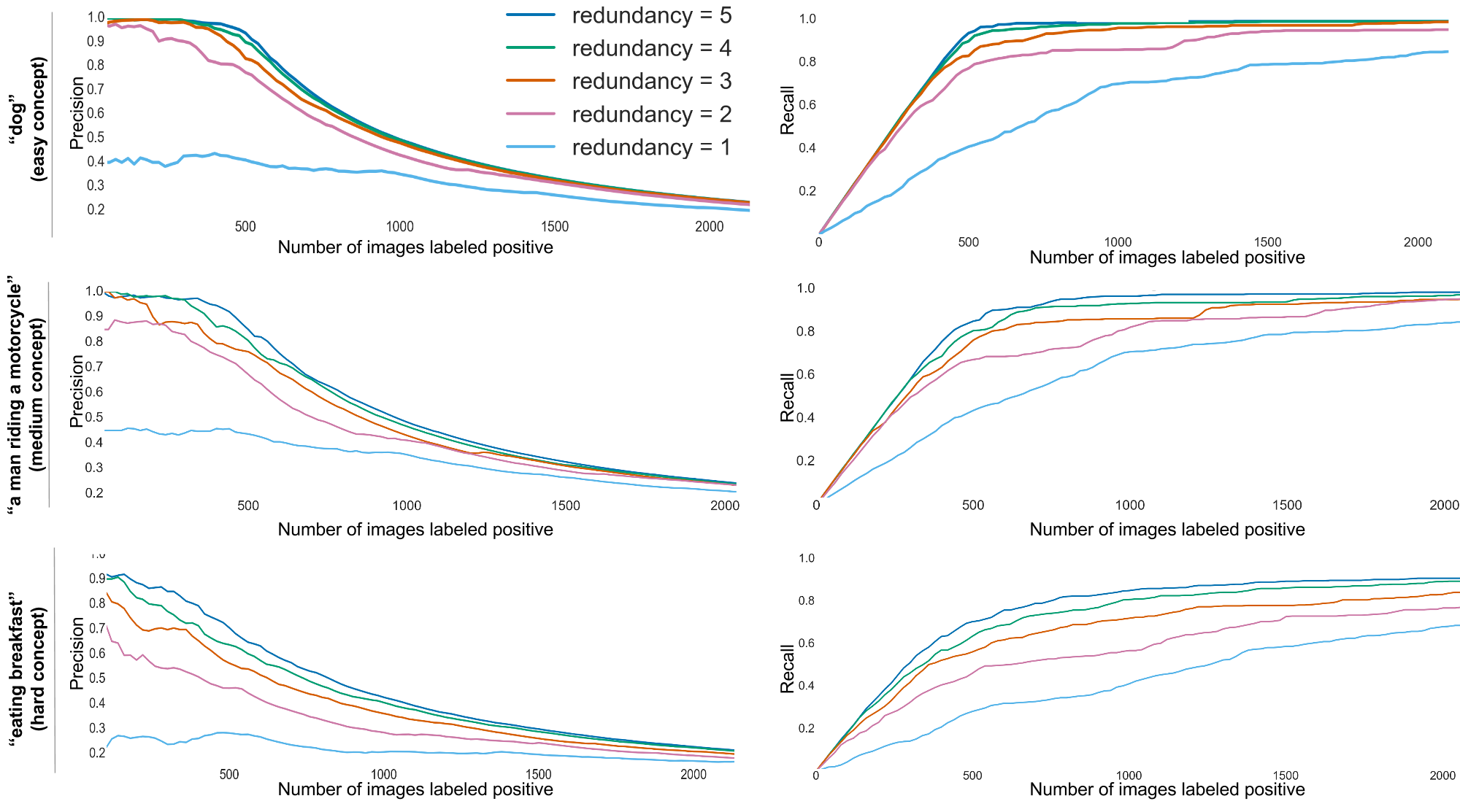}
\caption{We study the precision (left) and recall (right) curves for detecting ``dog'' (top), ``a person on a motorcycle'' (middle) and ``eating breakfast'' (bottom) images with a redundancy ranging from 1 to 5. There are 500 ground truth positive images in each experiment. We find that our technique works for simple as well as hard concepts.}
\label{fig:preception_cognition}
\end{figure}

\subsection{Calibration: Baseline Worker Reaction Time} 
Our technique hypothesizes that guiding workers to work quickly and make errors can lead to results that are faster yet with similar precision. We begin evaluating our technique by first studying worker reaction times as we vary the length of time for which each item is displayed. If worker reaction times have a low variance, we accurately model them. Existing work on RSVP estimated that humans usually react about 400ms after being presented with a cue~\cite{weichselgartner1987dynamics, reeves1986attention}. Similarly, the model human processor~\cite{card1983psychology} estimated that humans perceive, understand and react at least 240ms after a cue. We first measure worker reaction times, then analyze how frequently positive items can be displayed before workers are unable to react to them in time.

\runinhead{Method.} We recruited 1,000 workers on Amazon Mechanical Turk with 96\% approval rating and over 10,000 tasks submitted. Workers were asked to work on one task at a time. Each task contained a stream of 100 images of polka dot patterns of two different colors. Workers were asked to react by pressing the spacebar whenever they saw an image with polka dots of one of the two colors. Tasks could vary by two variables: the \textit{speed} at which images were displayed and the \textit{percentage} of the positively colored images. For a given task, we held the display speed constant. Across multiple tasks, we displayed images for 100ms to 500ms. We studied two variables: \textit{reaction time} and \textit{recall}. We measured the reaction time to the positive color across these speeds. To study recall (percentage of positively colored images detected by workers), we varied the ratio of positive images from 5\% to 95\%. We counted a keypress as a detection only if it occurred within 500ms of displaying a positively colored image.

\runinhead{Results.} Workers' reaction times corresponded well with estimates from previous studies. Workers tend to react an average of 378ms ($\sigma= 92$ms) after seeing a positive image. This consistency is an important result for our model because it assumes that workers have a consistent reaction delay.

As expected, recall is inversely proportional to the speed at which the images are shown. A worker is more likely to miss a positive image at very fast speeds. We also find that recall decreases as we increase the percentage of positive items in the task. To measure the effects of positive frequency on recall, we record the percentage threshold at which recall begins to drop significantly at different speeds and positive frequencies. From Figure~\ref{fig:recall_percentage}, at 100ms, we see that recall drops when the percentage of positive images is more than 35\%. As we increase the time for which an item is displayed, however, we notice that the drop in recall occurs at a much higher percentage. At 500ms, the recall drops at a threshold of 85\%. We thus infer that recall is inversely proportional to the \textit{rate} of positive stimuli and not to the percentage of positive images. From these results we conclude that at faster speeds, it is important to maintain a smaller percentage of positive images, while at slower speeds, the percentage of positive images has a lesser impact on recall. Quantitatively, to maintain a recall higher than 0.7, it is necessary to limit the frequency of positive cues to one every 400ms.

\subsection{Study 1: Image Verification}
In this study, we deploy our technique on image verification tasks and measure its speed relative to the conventional self-paced approach. Many crowdsourcing tasks in computer vision require verifying that a particular image contains a specific class or concept. We measure precision, recall and cost (in seconds) by the conventional approach and compare against our technique.

Some visual concepts are easier to detect than others. For example, detecting an image of a ``dog'' is a lot easier than detecting an image of ``a person riding a motorcycle'' or ``eating breakfast.'' While detecting a ``dog'' is a perceptual task, ``a person riding a motorcycle'' requires understanding of the interaction between the person and the motorcycle. Similarly, ``eating breakfast'' requires workers to fuse concepts of people eating a variety foods like eggs, cereal or pancakes. We test our technique on detecting three concepts: ``dog'' (easy concept), ``a person riding a motorcycle'' (medium concept) and ``eating breakfast'' (hard concept). In this study, we compare how workers fare on each of these three levels of concepts.

\runinhead{Method.} In this study, we compare the conventional approach with our technique on three (easy, medium and hard) concepts. We evaluate each of these comparisons using precision scores, recall scores and the speedup achieved. To test each of the three concepts, we labeled 10,000 images, where each concept had 500 examples. We divided the 10,000 images into streams of 100 images for each task. We paid workers \$0.17 to label a stream of 100 images (resulting in a wage of \$6 per hour~\cite{salehi2015we}). We hired over 1,000 workers for this study satisfying the same qualifications as the calibration task.

The conventional method of collecting binary labels is to present a crowd worker with a set of items. The worker proceeds to label each item, one at a time. Most datasets employ multiple workers to label each task because majority voting~\cite{snow2008cheap} has been shown to improve the quality of crowd annotations. These datasets usually use a redundancy of 3 to 5 workers~\cite{sheshadri2013square}. In all our experiments, we used a redundancy of 3 workers as our baseline. 

When launching tasks using our technique, we tuned the image display speed to 100ms. We used a redundancy of 5 workers when measuring precision and recall scores. To calculate \textit{speedup}, we compare the total worker time taken by all the 5 workers using our technique with the total worker time taken by the 3 workers using the conventional method. Additionally, we vary redundancy on all the concepts to from 1 to 10 workers to see its effects on precision and recall.

\runinhead{Results.} Self-paced workers take 1.70s on average to label each image with a concept in the conventional approach (Table~\ref{tab:precision_recall_speedup}). They are quicker at labeling the easy concept (1.50s per worker) while taking longer on the medium (1.70s) and hard (1.90s) concepts.

Using our technique, even with a redundancy of 5 workers, we achieve a speedup of 10.20$\times$ across all concepts. We achieve \textit{order of magnitude} speedups of $9.00\times$, $10.20\times$ and $11.40\times$ on the easy, medium and hard concepts. Overall, across all concepts, the precision and recall achieved by our technique is 0.97 and 0.81. Meanwhile the precision and recall of the conventional method is 0.97 and 0.96. We thus achieve the same precision as the conventional method. As expected, recall is lower because workers are not able to detect every single true positive example. As argued previously, lower recall can be an acceptable tradeoff when it is easy to find more unlabeled images.

Now, let's compare precision and recall scores between the three concepts. We show precision and recall scores in Figure~\ref{fig:preception_cognition} for the three concepts. Workers perform slightly better at finding ``dog'' images and find it the most difficult to detect the more challenging ``eating breakfast'' concept. With a redundancy of 5, the three concepts achieve a precision of 0.99, 0.98 and 0.90 respectively at a recall of 0.94, 0.83 and 0.74 (Table~\ref{tab:precision_recall_speedup}). The precision for these three concepts are identical to the conventional approach, while the recall scores are slightly lower. The recall for a more difficult cognitive concept (``eating breakfast'') is much lower, at 0.74, than for the other two concepts. More complex concepts usually tend to have a lot of contextual variance. For example, ``eating breakfast'' might include a person eating a ``banana,'' a ``bowl of cereal,'' ``waffles'' or ``eggs.'' We find that while some workers react to one variety of the concept (e.g.,\ ``bowl of cereal''), others react to another variety (e.g.,\ ``eggs''). 

\begin{figure}
\centering
\includegraphics[width=\linewidth]{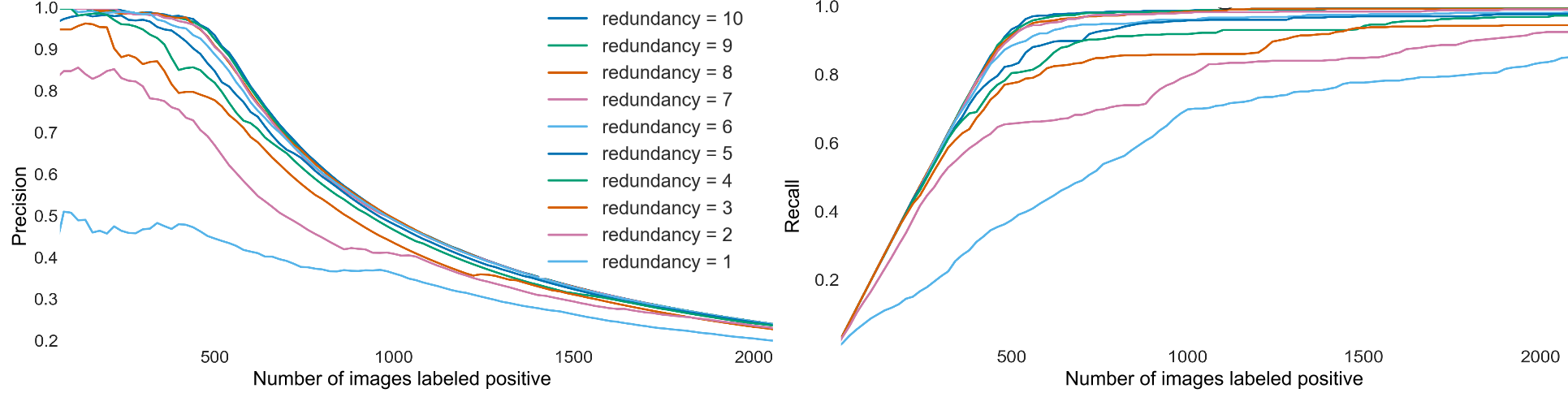}
\caption{We study the effects of redundancy on recall by plotting precision and recall curves for detecting ``a person on a motorcycle'' images with a redundancy ranging from 1 to 10. We see diminishing increases in precision and recall as we increase redundancy. We manage to achieve the same precision and recall scores as the conventional approach with a redundancy of 10 while still achieving a speedup of $5\times$.}
\label{fig:redundancy_10}
\end{figure}

When we increase the redundancy of workers to 10 (Figure~\ref{fig:redundancy_10}), our model is able to better approximate the positive images. We see diminishing increases in both recall and precision as redundancy increases. At a redundancy of 10, we increase recall to the same amount as the conventional approach (0.96), while maintaining a high precision (0.99) and still achieving a speedup of $5.1\times$.

We conclude from this study that our technique (with a redundancy of 5) can speed up image verification with easy, medium and hard concepts by an order of magnitude while still maintaining high precision. We also show that recall can be compensated by increasing redundancy.

\subsection{Study 2: Non-Visual Tasks}
So far, we have shown that rapid crowdsourcing can be used to collect image verification labels. We next test the technique on a variety of other common crowdsourcing tasks: sentiment analysis~\cite{pang2008opinion}, word similarity~\cite{snow2008cheap} and topic detection~\cite{LEWIS94d}.

\runinhead{Method.} In this study, we measure precision, recall and speedup achieved by our technique over the conventional approach. To determine the stream speed for each task, we followed the prescribed method of running trials and speeding up the stream until the model starts losing precision. For sentiment analysis, workers were shown a stream of tweets and asked to react whenever they saw a positive tweet. We displayed tweets at 250ms with a redundancy of 5 workers. For word similarity, workers were shown a word (e.g.,\ ``lad'') for which we wanted synonyms. They were then rapidly shown other words at 600ms and asked to react if they see a synonym (e.g.,\ ``boy''). Finally, for topic detection, we presented workers with a topic like ``housing'' or ``gas'' and presented articles of an average length of 105 words at a speed of 2s per article. They reacted whenever they saw an article containing the topic we were looking for. For all three of these tasks, we compare precision, recall and speed against the self-paced conventional approach with a redundancy of 3 workers. Every task, for both the conventional approach and our technique, contained 100 items. 

To measure the cognitive load on workers for labeling so many items at once, we ran the widely-used NASA Task Load Index (TLX)~\cite{colligan2015cognitive} on all tasks, including image verification. TLX measures the perceived workload of a task. We ran the survey on 100 workers who used the conventional approach and 100 workers who used our technique across all tasks.

\begin{figure}[h]
\centering
\includegraphics[width=\linewidth]{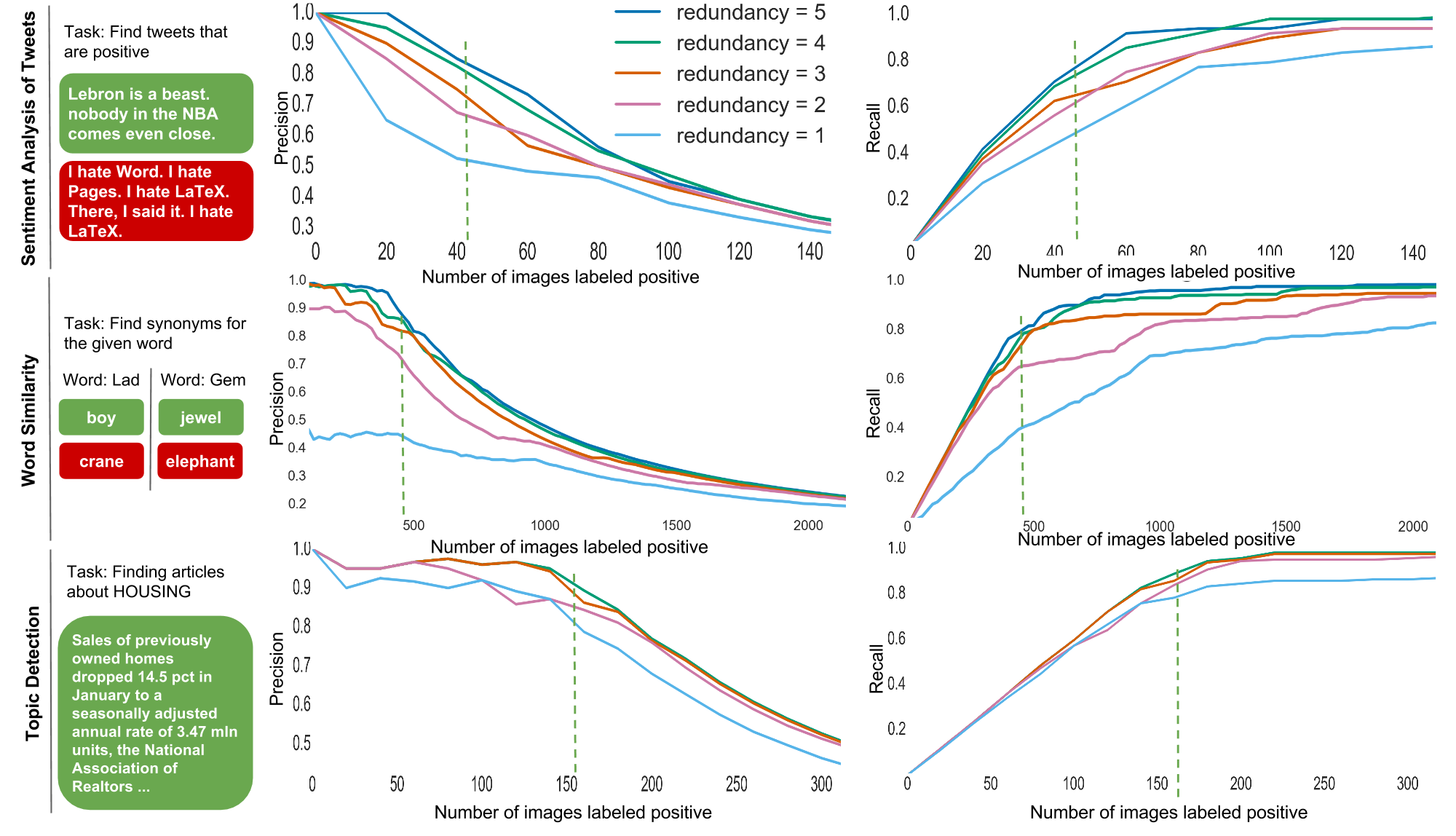}
\caption{Precision (left) and recall (right) curves for sentiment analysis (top), word similarity (middle) and topic detection (bottom) images with a redundancy ranging from 1 to 5. Vertical lines indicate the number of ground truth positive examples.}
\label{fig:other_tasks}
\end{figure}

\runinhead{Results.} We present our results in Table~\ref{tab:precision_recall_speedup} and Figure~\ref{fig:other_tasks}. For sentiment analysis, we find that workers in the conventional approach classify tweets in 4.25s. So, with a redundancy of 3 workers, the conventional approach would take 12.75s with a precision of 0.93. Using our method and a redundancy of 5 workers, we complete the task in 1250ms (250ms per worker per item) and 0.94 precision. Therefore, our technique achieves a speedup of $10.2\times$.

Likewise, for word similarity, workers take around 6.23s to complete the conventional task, while our technique succeeds at 600ms. We manage to capture a comparable precision of 0.88 using 5 workers against a precision of 0.89 in the conventional method with 3 workers. Since finding synonyms is a higher-level cognitive task, workers take longer to do word similarity tasks than image verification and sentiment analysis tasks. We manage a speedup of $6.23\times$.

Finally, for topic detection, workers spend significant time analyzing articles in the conventional setting (14.33s on average). With 3 workers, the conventional approach takes 43s. In comparison, our technique delegates 2s for each article. With a redundancy of only 2 workers, we achieve a precision of 0.95, similar to the 0.96 achieved by the conventional approach. The total worker time to label one article using our technique is 4s, a speedup of $10.75\times$.

The mean TLX workload for the control condition was 58.5 ($\sigma=9.3$), and 62.4 ($\sigma=18.5$) for our technique. Unexpectedly, the difference between conditions was not significant ($t(99)=-0.53, p=0.59$). The “temporal demand” scale item appeared to be elevated for our technique (61.1 vs. 70.0), but this difference was not significant ($t(99)=-0.76, p=0.45$). We conclude that our technique can be used to scale crowdsourcing on a variety of tasks without statistically increasing worker workload.

\subsection{Study 3: Multi-class Classification}
In this study, we extend our technique from binary to multi-class classification to capture an even larger set of crowdsourcing tasks. We use our technique to create a dataset where each image is classified into one category (``people,'' ``dog,'' ``horse,'' ``cat,'' etc.). We compare our technique with a conventional technique~\cite{deng2009imagenet} that collects binary labels for each image for every single possible class.

\runinhead{Method.} Our aim is to classify a dataset of 2,000 images with 10 categories where each category contains between 100 to 250 examples. We compared three methods of multi-class classification: (1) a \textit{naive approach} that collected 10 binary labels (one for each class) for each image, (2) a \textit{baseline approach} that used our interface and classified images one class (chosen randomly) at a time, and (3) a \textit{class-optimized approach} that used our interface to classify images starting from the class with the most examples. When using our interface, we broke tasks into streams of 100 images displayed for 100ms each. We used a redundancy of 3 workers for the conventional interface and 5 workers for our interface. We calculated the precision and recall scores across each of these three methods as well as the cost (in seconds) of each method.

\runinhead{Results.} (1) In the \textit{naive approach}, we need to collect 20,000 binary labels that take 1.7s each. With 5 workers, this takes 102,000s (\$170 at a wage rate of \$6/hr) with an average precision of $0.99$ and recall of $0.95$. (2) Using the \textit{baseline approach}, it takes 12,342s (\$20.57) with an average precision of $0.98$ and recall of $0.83$. This shows that the baseline approach achieves a speedup of $8.26\times$ when compared with the naive approach. (3) Finally, the \textit{class-optimized approach} is able to detect the most common class first and hence reduces the number of times an image is sent through our interface. It takes 11,700s (\$19.50) with an average precision of $0.98$ and recall of $0.83$. The class-optimized approach achieves a speedup of $8.7\times$ when compared to the naive approach. While the speedup between the baseline and the class-optimized methods is small, it would be increased on a larger dataset with more classes.

\subsection{Application: Building ImageNet}
Our method can be combined with existing techniques~\cite{deng2014scalable, song2011contextualizing, parkash2012attributes, biswas2013simultaneous} that optimize binary verification and multi-class classification by preprocessing data or using active learning. One such method~\cite{deng2014scalable} annotated ImageNet (a popular large dataset for image classification) effectively with a useful insight: they realized that its classes could be grouped together into higher semantic concepts. For example, ``dog,'' ``rabbit'' and ``cat'' could be grouped into the concept ``animal.'' By utilizing the hierarchy of labels that is specific to this task, they were able to preprocess and reduce the number of labels needed to classify all images. As a case study, we combine our technique with their insight and evaluate the speedup in collecting a subset of ImageNet.

\runinhead{Method.} We focused on a subset of the dataset with 20,000 images and classified them into 200 classes. We conducted this case study by comparing three ways of collecting labels: (1) The naive approach asked 200 binary questions for each image in the subset, where each question asked if the image belonged to one of the 200 classes. We used a redundancy of 3 workers for this task. (2) The optimal-labeling method used the insight to reduce the number of labels by utilizing the hierarchy of image classes.  (3) The combined approach used our technique for multi-class classification combined with the hierarchy insight to reduce the number of labels collected. We used a redundancy of 5 workers for this technique with tasks of 100 images displayed at 250ms.

\runinhead{Results.} (1) Using the naive approach, this would result in asking 4 million binary verification questions. Given that each binary label takes 1.7s (Table~\ref{tab:precision_recall_speedup}), we estimate that the total time to label the entire dataset would take 6.8 million seconds (\$11,333 at a wage rate of \$6/hr). (2) The optimal-labeling method is estimated to take 1.13 million seconds (\$1,888)~\cite{deng2014scalable}. (3) Combining the hierarchical questions with our interface, we annotate the subset in 136,800s (\$228).  We achieve a precision of $0.97$ with a recall of $0.82$. By combining our $8\times$ speedup with the $6\times$ speedup from intelligent question selection, we achieve a $50\times$ speedup in total.

\subsection{Discussion}

\runinhead{Absence of Concepts.} 
We focused our technique on positively identifying concepts. We then also test its effectiveness at classifying the absence of a concept. Instead of asking workers to react when they see a ``dog,'' if we ask them to react when they do \textit{not} see a ``dog,'' our technique performs poorly. At $100$ms, we find that workers achieve a recall of only $0.31$, which is much lower than a recall of $0.94$ when detecting the presence of ``dog''s. To improve recall to $0.90$, we must slow down the feed to $500$ms. Our technique achieves a speedup of $2\times$ with this speed. We conclude that our technique performs poorly for anomaly detection tasks, where the presence of a concept is common but its absence, an anomaly, is rare. More generally, this exercise suggests that some cognitive tasks are less robust to rapid judgments. Preattentive processing can help us find ``dog''s, but ensuring that there is no ``dog'' requires a linear scan of the entire image.

\runinhead{Typicality.} 
To better understand the active mechanism behind our technique, we turn to concept typicality. A recent study~\cite{iordan2015basic} used fMRIs to measure humans' recognition speed for different object categories, finding that images of most typical examplars from a class were recognized faster than the least typical categories. They calculated typicality scores for a set of image classes based on how quickly humans recognized them.
In our image verification task, $72\%$ of false negatives were also atypical. 
Not detecting atypical images might lead to the curation of image datasets that are biased towards more common categories. For example, when curating a dataset of dogs, our technique would be more likely to find usual breeds like ``dalmatians'' and ``labradors'' and miss rare breeds like ``romagnolos'' and ``otterhounds.'' More generally, this approach may amplify biases and minimize clarity on edge cases. Slowing down the feed reduces atypical false negatives, resulting in a smaller speedup but with a higher recall for atypical images. 

\runinhead{Conclusion.}
We have suggested that crowdsourcing can speed up labeling by encouraging a small amount of error rather than forcing workers to avoid it. We introduce a rapid slideshow interface where items are shown too quickly for workers to get all items correct. We algorithmically model worker errors and recover their intended labels. This interface can be used for binary verification tasks like image verification, sentiment analysis, word similarity and topic detection, achieving speedups of $10.2\times$, $10.2\times$, $6.23\times$ and $10.75\times$ respectively. It can also extend to multi-class classification and achieve a speedup of $8.26\times$. Our approach is only one possible interface instantiation of the concept of encouraging some error; we suggest that future work may investigate many others. Speeding up crowdsourcing enables us to build larger datasets to empower scientific insights and industry practice. For many labeling goals, this technique can be used to construct datasets that are an order of magnitude larger without increasing cost.

\section{Data acquisition through social interactions}
\label{sec:3}
Modern supervised machine learing (ML) systems in domains such as computer vision are reliant on mountains of human-labeled training data. These labeled images, for example the fourteen million images in ImageNet~\cite{deng2009imagenet}, require basic human knowledge such as whether an image contains a chair. Unfortunately, this knowledge is both so simple that it is extremely tedious for humans to label, and also so tacit that the human annotators are required. In response, crowdsourcing efforts often recruit volunteers to help create labels via intrinsic interest, curiosity or gamification~\cite{lintott2008galaxy,law2016curiosity,willis2017crowdcurio,von2004labeling}. 

The general approach of these crowdsourcing efforts is to focus on \textit{what} to ask each contributor. Specifically, from a large set of possible tasks, many systems formalize an approach to route or recommend tasks to specific contributors~\cite{GEIGER20143,AAAI148425,AAAIW114005,Difallah:2013:PTM:2488388.2488421}.
Unfortunately, many of these volunteer efforts are restricted to labels for which contributions can be motivated, leaving incomplete any task that is uninteresting to contributors~\cite{reich2012state,hill2013almost,healy2003ecology,warncke2015misalignment}.

\begin{figure}[h]
\centering
\sidecaption
\includegraphics[width=0.65\linewidth]{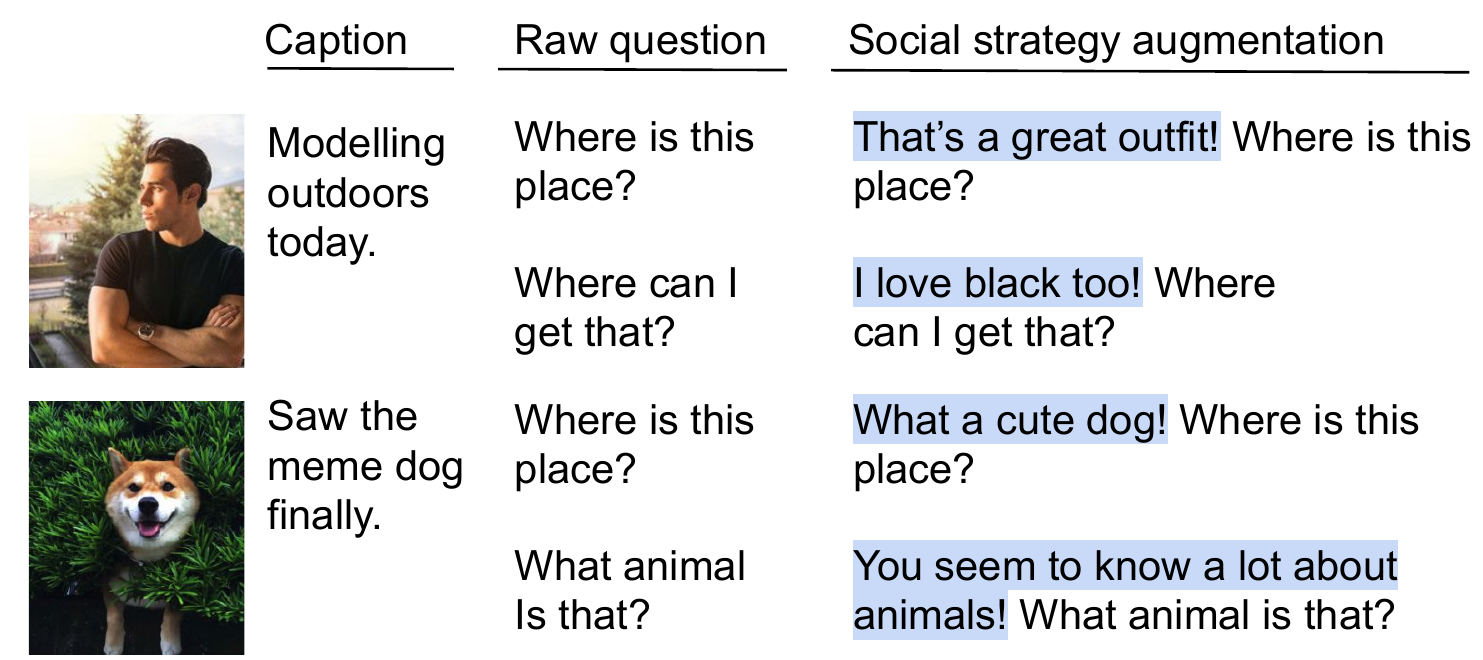}
\caption{We introduce an approach that increases crowdsourcing participation rates by learning to augment requests with image- and text-relevant question asking strategies drawn from social psychology. Given a social media image post and a question, our approach selects a strategy and generates a natural language phrase to augment the question.}
\label{fig:pull_figure}
\end{figure}

\begin{figure*}[t]
\centering
    {\includegraphics[width=\textwidth]{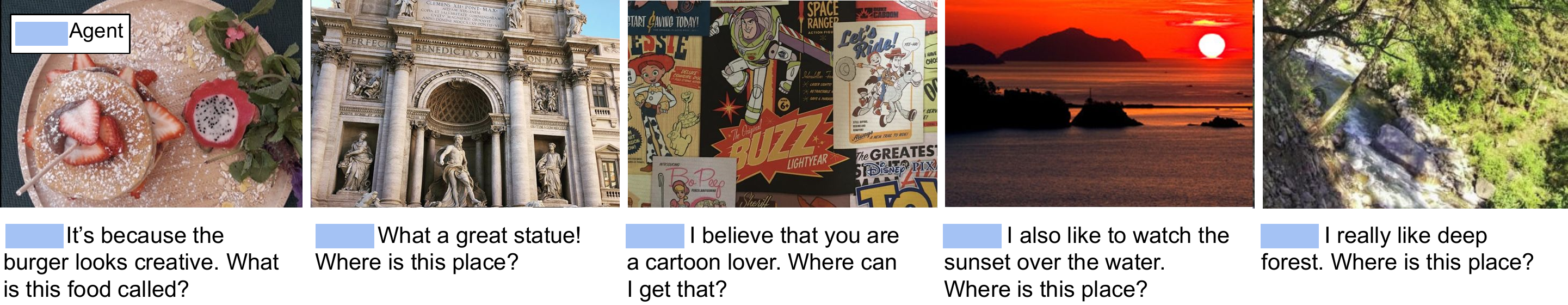}}
    \caption{Our agent chooses appropriate social strategies and contextualizes questions to maximize crowdsourcing participation. 
    }
    \label{fig:CommentExamples}
\end{figure*}

Our paper specifically studies an instantiation of this common ailment in the context of visual question answering (VQA). VQA generalizes numerous computer vision tasks, including object detection~\cite{deng2009imagenet}, relationship prediction~\cite{lu2016visual}, and action prediction~\cite{niebles2008unsupervised}. Progress in VQA supports the development of many human-computer interaction systems, including VizWiz~\cite{bigham2010vizwiz}, TapTapSee, BeMyEyes, and CamFind\footnote{Applications can be found at \url{https://taptapsee.com/}, https://www.bemyeyes.com/, and \url{https://camfindapp.com/}}. VQA is a data-hungry machine learning task that is challenging to motivate contributors. Existing VQA crowdsourcing strategies have suggested using social media to incentivize online participants to answer visual questions for assistive users~\cite{bigham2010vizwiz,Brady:2015:GRS:2702123.2702329}, but many such questions remain unanswered~\cite{brady2013investigating}.

To meet the needs of modern ML systems, we argue that crowdsourcing systems can automatically generate plans not just for \textit{what} to ask about, but also for \textit{how} to make that request.
Social psychology and social computing research have made clear that how a request is structured can have substantial effects on resulting contribution rates~\cite{kraut2011encouraging,Yang:2017:PTG:3171581.3134749}.
However, while it is feasible to manually design a single request such as one email message to all users in an online community, or one motivational message on all web pages on Wikipedia, in real life (as in VQA) there exist a wide variety of situations that must each be approached differently.
Supporting this variety in \textit{how} a request is made has remained out of reach; in this paper, we contribute algorithms to achieve it.

Consider, for example, that we are building a dataset of images with their tagged geolocations (Figure~\ref{fig:pull_figure}). When we encounter an image of a person wearing a black shirt next to a beautiful scenery, existing machine learning systems can generate questions such as ``where is this place?''. However, prior work reports that such requests seem mechanical, resulting in lower response rates~\cite{brady2013investigating}.  In our approach, requests might be augmented by \texttt{content compliment} strategies~\cite{CialdiniInfluence} reactive to the image content, such as ``What a great statue!'' or ``That's a beautiful building!'', or by \texttt{interest matching} strategies~\cite{CialdiniPreSuasion} reactive to the image content, such as ``I love visiting statues!'' or ``I love seeing old buildings!''

Augmenting requests with social strategies requires (1)~defining a set of possible social strategies, (2)~developing a method to generate content for each strategy conditioned on an image, and (3)~choosing the appropriate strategy to maximize response conditioned on the user and their post. In this paper, we tackle these three challenges. First, we adopt a set of social strategies that social psychologists have demonstrated to be successful in human-human communication~\cite{CialdiniPreSuasion,CialdiniInfluence,Langer,taylor,HoffmanML}. While our set is not exhaustive, it represents a diverse list of strategies --- some that augment questions conditioned on the image and others conditioned on the user's language. While previous work has explored the use of ML models to generate image-conditioned natural language fragments, for generating captions and questions, ours is the first method that employs these techniques to generate strategies that increase worker participation.

To test the efficacy of our approach, we deploy our system on Instagram, a social media image-sharing platform. We collect datasets and develop machine learning-based models that use a convolutional neural network (CNN) to encode the image contents and a long short-term memory network (LSTM) to generate each social strategy across a large set of different kinds of images. We compare our ML strategies against baseline rule-based strategies using linguistic features extracted from the user's post~\cite{li2010contextual}. We show a sample of augmented questions in Figure~\ref{fig:CommentExamples}. We find that choosing appropriate strategies and augmenting requests leads to a significant absolute participation increase of $42.36\%$ over no strategy when using ML strategies and a $14.78\%$ increase when using rule-based strategies. We also find that no specific strategy is the universal best choice, implying that knowing when to use a strategy is important. While we specifically focus on VQA and Instagram, our approach generalizes to other crowdsourcing systems that support language-based interaction with contributors.


\subsection{Related work}
Our work is motivated by research in crowdsourcing, peer production and social computing that increase contributors' levels of intrinsic motivation. We thread this work together with advances in natural language generation technologies to contribute generative algorithms that  modulating the form of the requests to increase contribution rates.

\runinhead{Crowdsourcing strategies.} 
The HCI community has investigated different ways to incentivise people to participate in data-labeling tasks~\cite{hill2013almost,healy2003ecology,reich2012state}. Designing for curiosity, for example, increases crowdsourcing participation~\cite{law2016curiosity}. Citizen science projects like GalaxyZoo mobilize volunteers by motivating them to work on a domain that aligns with their interests~\cite{lintott2008galaxy}. Unlike the tasks typically explored by such methods, image-labeling is not typically an intrinsically motivated task, and is instead completed by paid ghost work~\cite{ghostwork}. To improve image-labeling, the ESP Game harnessed game design to solve annotation tasks as by-products of entertainment activities~\cite{vonahn}. However, games result in limited kinds of labels, and need to be designed specifically to attain certain types of labels. Instead, we ask directed questions through conversations to label data and use social strategies to motive participation.

\runinhead{Interaction through conversations.}
The use of natural language as a medium for interaction has galvanized many systems~\cite{huang2018evorus,lasecki2013chorus}. Natural language has been proposed as a medium to gather new data from online participants~\cite{bigham2010vizwiz} or guide users through workflows~\cite{fast2018iris}. Conversational agents have also been deployed through products like  Apple's Siri, Amazon's Echo, and Microsoft's Cortana. Substantial effort has been placed on teaching people how to talk to  such assistants. Noticing this limitation, more robust crowd-powered conversational systems have been created by hiring professionals, as in the case of Facebook M~\cite{hempel2015facebook}, or crowd workers~\cite{lasecki2013chorus,bohus2009ravenclaw}. Unlike these approaches where people have a goal and invoke a passive conversational agent, we build active agents reach out to people with questions that increase humans participation.

\runinhead{Social interaction with machines.}
To design an agent capable of eliciting a user's help, we need to understand how a user views the interaction. The Media Equation proposes that people adhere to similar social norms in their interactions with computers as they do in interactions with other people~\cite{reeves1996media}. It shoes that agents that seem more human-like, in terms of behaviour and gestures, provoke users to treat them similar to a person~\cite{Cassell1999ThePO,Cerrato2002DifferentWO,Nass:2007:WSV:1197573}. Consistent with these observations, prior work has also shown that people are more likely to resolve misunderstandings with more human-like agents~\cite{CORTI2016431}. This leads us to question whether a human-like conversational agent can encourage more online participation from online contributors. 
Prior work on interactions with machines investigates social norms that a machine can mimic in a binary capacity --- either it respects the norm correctly or violates it with negligence~\cite{sardar,Chidambaram}. Instead, we project social interaction on a spectrum --- some social strategies are more successful than others in a given context --- and learn a selection strategy that maximizes participation.

\runinhead{Structuring requests to enhance motivation.} 
There have been many proposed social strategies to enhance the motivation to contribute in online communities~\cite{kraut2011encouraging}. For example, asking a specific question rather than
making a statement or asking an open-ended question increases the likelihood of getting a
response~\cite{burkemembership}. Requests succeed significantly more often when contributors are addressed by name~\cite{markey2000bystander}. Emergencies receive more responses than requests without time constraints~\cite{darley1968bystander}. Prior work has shown that factors that increase the contributor's affinity for the requester increase the persuasive power of the message on online crowdfunding sites~\cite{Yang:2017:PTG:3171581.3134749}. It has also been observed that different behaviour elicits different kind of support from online support groups with self disclosure eliciting emotional support and questioning resulting in informational support~\cite{info:doi/10.2196/jmir.3558}. The severity of the outcome of responding to a request can also influence motivation~\cite{chaiken1989heuristic}. 
Our work incorporates some of these established social strategies and leverages  language generation algorithms to build an agent that can deploy them across a wide variety of different requests. 

\begin{figure}[t]
\centering
\includegraphics[width=\linewidth]{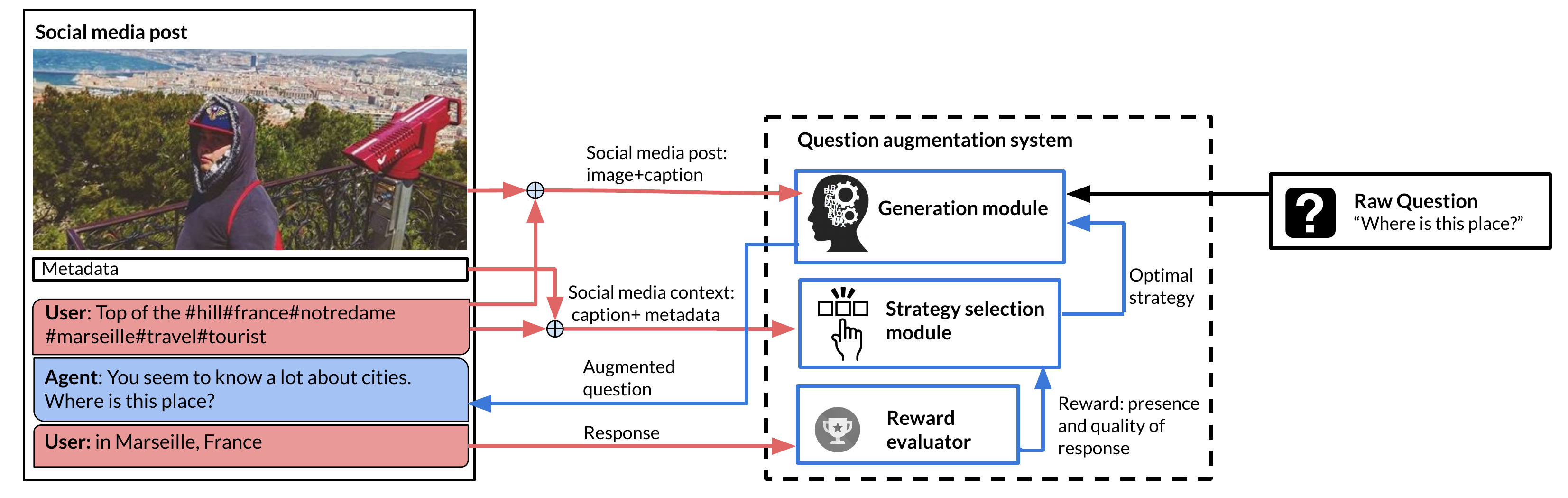}
\caption{Given a social media post and a question we want to ask, we augment the question with a social strategy. Our system contains two components. First, a selection component featurizes the post and user and chooses a social strategy. Second, a generation component creates a natural language augmentation for the question given the image and the chosen strategy. The contributor's response or silence is used to generate a feedback reward for the selection module.}
\label{fig:system}
\end{figure}

\subsection{Social strategies}
The goal of our system is to draw on theories of how people ask other people for help and favors, then learn how to emulate those strategies. Drawing on prior work, we sampled a diverse set of nine social strategies. While the set of nine social strategies we explore are not an exhaustive set, we believe it represents a wide enough range of possible strategies to demonstrate the method and effects of teaching social strategies to machines. The social strategies we explore are:

\begin{enumerate}
    \item \texttt{Content compliment}: Compliment the image or an object in the image before asking the question. This increases the liking between the agent and the contributor, making them more likely to reciprocate with the request~\cite{CialdiniInfluence}. 
    \item \texttt{Expertise compliment}: Compliment the knowledge of the contributor who posted the image. This commits the contributor as an ``expert'', resulting in a thoughtful response~\cite{CialdiniInfluence}. 
    \item \texttt{Interest matching}: Show interest in the topic of the contributor's post. This creates a sense of unity between the agent and contributor~\cite{CialdiniPreSuasion}. 
    \item \texttt{Valence matching}: Match the valence of the contributor based on their image's caption. People evolved to act kindly to others who exhibit behaviors from a similar culture~\cite{taylor}.
    \item \texttt{Answer attempt}: Guess an answer and ask for a validation. Recognizing whether a shown answer is correct or not is cognitively an easier task for the listener than recalling the correct answer~\cite{gillund1984retrieval}.
    \item \texttt{Time scarcity}: Specify an arbitrary deadline for the response. People are more likely to act if the opportunity is deemed to expire, even if they neither need nor want the opportunity~\cite{CialdiniInfluence}.
    \item \texttt{Help request}: Explicitly request the contributor's help. People are naturally inclined to help others when they are asked and able to do so~\cite{HoffmanML}.
    \item \texttt{Logical justification}: Give a logical reason for asking the question to persuade the contributor at a cognitive level~\cite{Langer}.
    \item \texttt{Random justification}: Give a random reason for asking the question. People are more likely to help if a justification is provided, even if it does not actually entail the request~\cite{Langer}.
\end{enumerate}

\subsection{System Design}
In this section, we describe our approach for augmenting requests with social strategies. Our approach is divided into two components: generation and selection. A high-level system diagram is depicted in Figure~\ref{fig:system}. Given a social media post, we featurize the post metadata, question, and caption, then send them to the selection component. The main goal of the selection component is to choose an effective social strategy to use for the given post. This strategy, along with a generated question to ask~\cite{krishna2019information}, and the social media post are sent to the generation component, which augments the question by generating a natural language phrase for the chosen social strategy. The augmented request is then shared with the contributor. The selection module gathers feedback, positive if the contributor responds in an informative manner. Uninformative responses or no response are counted as a negative feedback.

\begin{figure}[h]
\centering
\includegraphics[width=\linewidth]{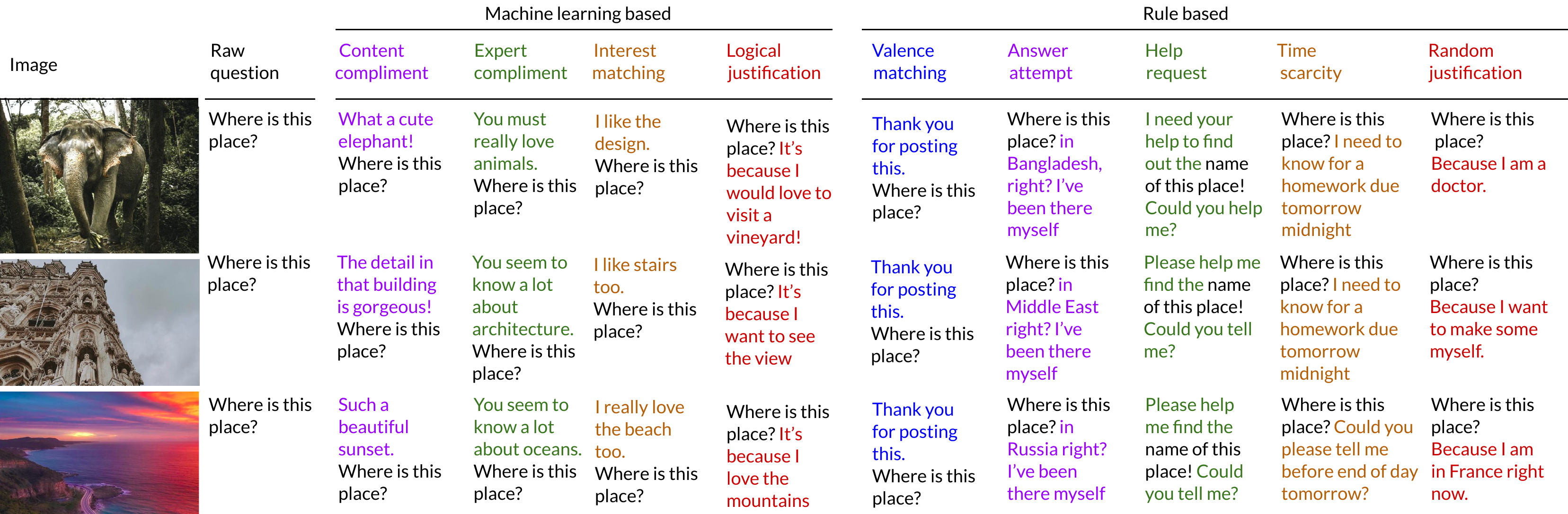}
\caption{Example augmentations generated by each of our social strategies.}
\label{fig:examples_strategies}
\end{figure}

\subsubruninhead{Selection: Choosing a social strategy}
We model our selection component as a contextual bandit. Contextual bandits are a common reinforcement learning technique for efficiently exploring different options and exploiting the best choices over time, generalizing from previous trials to uncommonly observed situations~\cite{li2010contextual}. The component receives a feature vector and outputs its choice of an arm (option) that it expects to result in the highest expected reward. 

Each social media post is represented as a feature vector that encodes information about the user, the post, and the caption. User features include- number of posts the user has posted, number of followers, number of accounts the user is following, number of other users tagged in their posts, filters and AR effects the user uses frequently on the platform, user's engagement with videos, whether the user is a verified business or an influencer, user's  privacy settings, the engagement with Instagram features such as highlight reels and resharing, and sentiment analysis on their biography. Post features include the number of users who like the post and the number of users who commented on the post. User and post features are drawn from Instagram's API and featurized as bag of words or one-hot vectors. Lastly, caption features are extracted from sentiment using Vader~\cite{hutto2014vader}, and the hashtags extracted using regular expressions. 

We train a contextual bandit model to choose a social strategy given the extracted features, conditioned on the success of each social strategy used on similar social media posts in the past.  The arms that the contextual bandit considers represent each of the nine social strategies that the system can use. If a chosen social strategy receives a response, we parse and check if the response contains an answer~\cite{devlin2018bert}. If so, the model receives a positive reward for choosing the social strategy. If a chosen social strategy does not receive a response, or if the response does not contain an answer, the model receives a negative reward.

Our implementation of contextual bandit uses the adaptive greedy algorithm for balancing the trade-off between exploration and exploitation. During training, the algorithm chooses an option that the model associates with a high uncertainty of reward. If there is no option with a high uncertainty, the algorithm chooses a random option to explore. The threshold for uncertainty decreases as the model is exposed to more data. During inference, the model predicts the social strategy with highest expected reward~\cite{zhang2004solving}.

\subsubruninhead{Generation: Augmenting questions}
The generation component receives the social media post (an image and a caption) and a raw question automatically generated by existing visual question generation algorithms (e.g., ``Where is this place?''). It produces a natural language contextualization of the question using one of the nine social strategies chosen by the selection component.

We build nine independent natural language generation systems that each receive a social media post as input and produce a comment using the corresponding social strategy as output. Four of the social strategies require knowledge about the content of images, and are implemented using machine learning-based models. These strategies cannot be templatized, as there is substantial variation in the kinds of images found online and the approaches much be personalized to the content of the image. We use the other five social strategies as baseline strategies that only require knowledge about the speaking style of the social media user, and are implemented as rule-based expert systems in conjunction with natural language processing techniques. We discuss these two types of models below.

\runinhead{Machine learning-based social strategies.}
To generate sentences specific to the image of each post, we train one machine learning model for each of the four social strategies that require knowledge about the image: \texttt{expert compliment}, \texttt{content compliment}, \texttt{interest matching}, and \texttt{logical justification}. 

We build a dataset of $10$k social media posts alongside examples of questions that use each of the four social social strategies, with the help of crowd workers on Amazon Mechanical Turk. This process results in a dataset of $40$k questions, each with social strategy augmentations. The posts are randomly selected by polling Instagram for images with one of the top $100$ most popular hashtags on Instagram and filter for those that refer to visual content, such as $\#$animal, $\#$travel, $\#$shopping, $\#$food, etc. Crowdworkers are designated to one of the four strategy categories and trained using examples and a qualifying task, which we manually evaluate. Each task contains $10$ social media posts (images and captions) and the generated questions. Workers are asked to submit a natural language sentence that can be pre- and post-pended to the question while adhering to the social strategy they are trained to emulate. The workers are paid a compensation that is equivalent to $\$12$ an hour for their work.\footnote{The dataset of social media posts and social strategies for training the reinforcement learning model, as well as the trained contextual bandit model, is publicly available at \url{http://cs.stanford.edu/people/ranjaykrishna/socialstrategies}.}

We adopt a traditional image-to-sequence machine learning model to generate the sentence for each strategy. Each model encodes the social media image using a convolutional neural network (CNN)~\cite{NIPS2012_4824} and generates a social strategy sentence, conditioned on image features, using a long short term memory (LSTM) network~\cite{hochreiter1997long}. We train each model using the dataset of $10$k posts dedicated to its assigned strategy using stochastic gradient descent with a learning rate of $1e-3$ for $15$ epochs.

\begin{figure}[t]
\centering
\includegraphics[width=\linewidth]{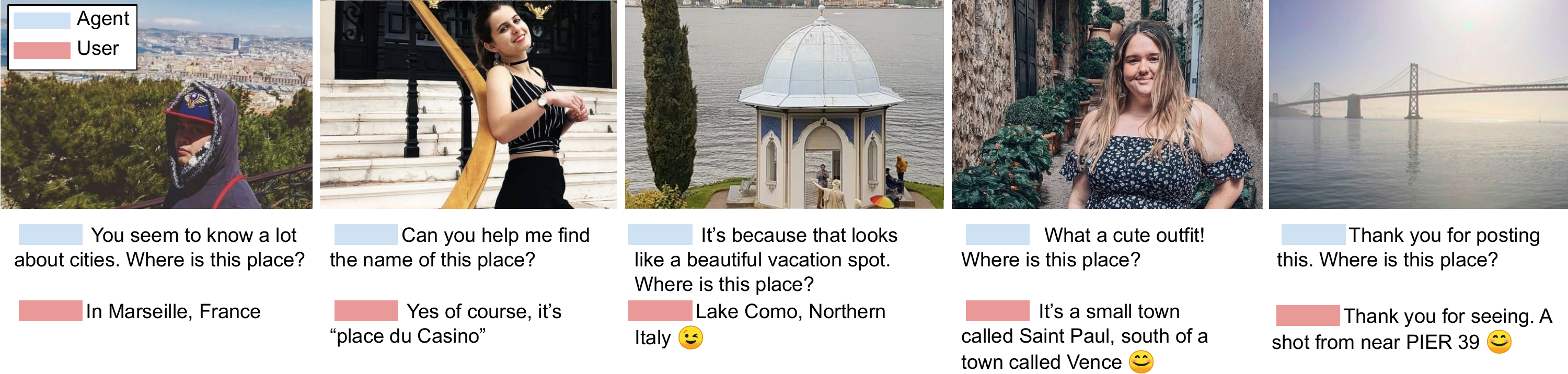}
\caption{Example responses to \texttt{expertise compliment}, \texttt{help request}, \texttt{logical justification}, \texttt{content compliment} and \texttt{valence matching} in the travel domain.} 
\label{fig:ESResponse2}
\end{figure}

\subsubruninhead{Baseline rule-based social strategies.}
To generate social strategy sentences that are relevant to the caption of each social media post, we create a rule-based expert system for each of the five social strategies: \texttt{valence matching}, \texttt{answer attempt}, \texttt{help request}, \texttt{time scarcity}, and \texttt{random justification}. While these algorithms use statistical machine learning approaches for natural language processing, we call them rule-based systems to clarify that the generation, itself, is a deterministic process unlike the sentences generated by the LSTM networks.

\texttt{Valence matching} detects the emotional valence of the caption through punctuation parsing and sentiment analysis using an implementation of the Vader algorithm~\cite{hutto2014vader}. The algorithm generates a sentence with emotional valence that is approximately equal to valence of the caption by matching type and number of punctuations and adding appropriate exclamations like ``Wow!'' or ``Aw''. 

\texttt{Answer attempt} guesses a probable answer for the input post based on the raw question and hashtags of the post. To guess a probable answer, we manually curate a set of likely answers for problem domains and words from caption and randomly choose one from the set. For example, when asking where we could buy the same item on a post that references the word ``jean'' in the ``$\#$shopping'' domain, the set of probable answers are a list of brands that sell jeans to consumers. Deployments of this strategy does not have to rely on a curated list and can instead use existing answering models~\cite{antol2015vqa}.

\texttt{Help request} augments the agent's question with variations of words and sentence structures that humans use to request help from one another. \texttt{Time scarcity} augments the agent's question with variations of a sentence that requests the answer to be provided within $24$ hours. \texttt{Random justification} augments the agent's question with a justification that is chosen irrespective of the social media post. Specifically, we store a list of justification sentences generated from the logical justification system for other posts, and retrieve one at random. Figure~\ref{fig:examples_strategies} visualizes example augmentations generated by each of our nine strategies, conditioned on the post.

\subsection{Experiments}
We evaluate the utility of augmenting questions with social strategies through a real-world deploying on Instagram. Our aim is to increase online crowdsourcing participation from Instagram users when we ask them questions about their image contents. We begin our experiments by first describing the experimental setup, the metrics used, the baselines, and strategies surveyed. Next, we study how generated social strategies impact participation. Finally, we study the importance of selecting the correct social strategy. 

\subsubruninhead{Experimental setup}
We poll images from Instagram, featurize the post, select a social strategy, and generate the question augmentation. We post the augmented question and wait for a response.

\runinhead{Images and raw questions.} We source images from Instagram across $4$ domains: travel, animals, shopping and food. Images from each domain are polled by searching for posts with hashtags: $\#$travel, $\#$animals, $\#$shopping, and $\#$food. Images in these four domains consitute an upper bound of $7.06\%$ of all images posted with one of the top $100$ popular hashtags that represent visual content.  Since we are studying the impact of using different social strategies by directly interacting with real users on Instagram, we can not post multiple questions, each augmented with a different strategy, to the same image post. Ideally, in online crowdsourcing deployments, the raw questions generated would be conditioned on the post or image. In our case, however, we use only one question per domain so that all users are exposed to the same basic question. For each domain, we hold the raw question constant. For example, ``Where is this place?'' for travel, ``What animal is that?'' for animals, ``Where can I get that?'' for shopping, and ``What is this food?'' for food.

\runinhead{Metrics.} To measure the improvements in crowdsourcing participation, we report the percentage of informative responses.  After a question is posted on Instagram, we wait $24$ hours to check if a response was received. If the question results in no response or if the response doesn't answer the question or the user appears confused (e.g.~``huh?'' or ``I don't understand''), the interaction is not counted as an informative response. To verify if a response is informative, we send all responses to Amazon Mechanical Turk (AMT) workers to report whether the question was actually answered with gold standard responses to guarantee quality.

\begin{table}[t]
\centering
\begin{tabular}{lll}
                                  & Source domain (\%) & Target domain (\%) \\ \hline
Expertise compliment              & \textbf{72.90}                                                         & 29.55                                                        \\
Content compliment                & 59.11                                                        & 68.96                                                        \\
Interest matching                 & 45.31                                                        & \textbf{85.38}                                                        \\
Logical justification             & 55.17                                                        & 19.7                                                         \\
Answer attempt                    & 41.37                                                        & 42.69                                                        \\
Help request                      & 31.52                                                        & 32.84                                                        \\
Valence matching                  & 37.43                                                        & 36.12                                                        \\
Time scarcity                     & 24.63                                                        & 26.27                                                        \\
Random justification              & 17.73                                                        & 32.84                                                        \\ \hline
                                  
ML based strategies & \textbf{58.12}                                                        & \textbf{50.89}                                                        \\
Rule based strategies             & 30.54                                                        & 34.15                                                        \\
No strategy                       & 15.76                                                        & 13.13 \\\hline
\end{tabular}
\label{tab:crossdomain}
\caption{Response rates achieved by different strategies on posts in the source and target domains. The bottom of the table shows a comparison between average performance of ML based strategies, average performance of rule-based strategies and baseline un-augmented questions }
\end{table}

\begin{figure}[t]
\centering
\includegraphics[width=0.85\linewidth]{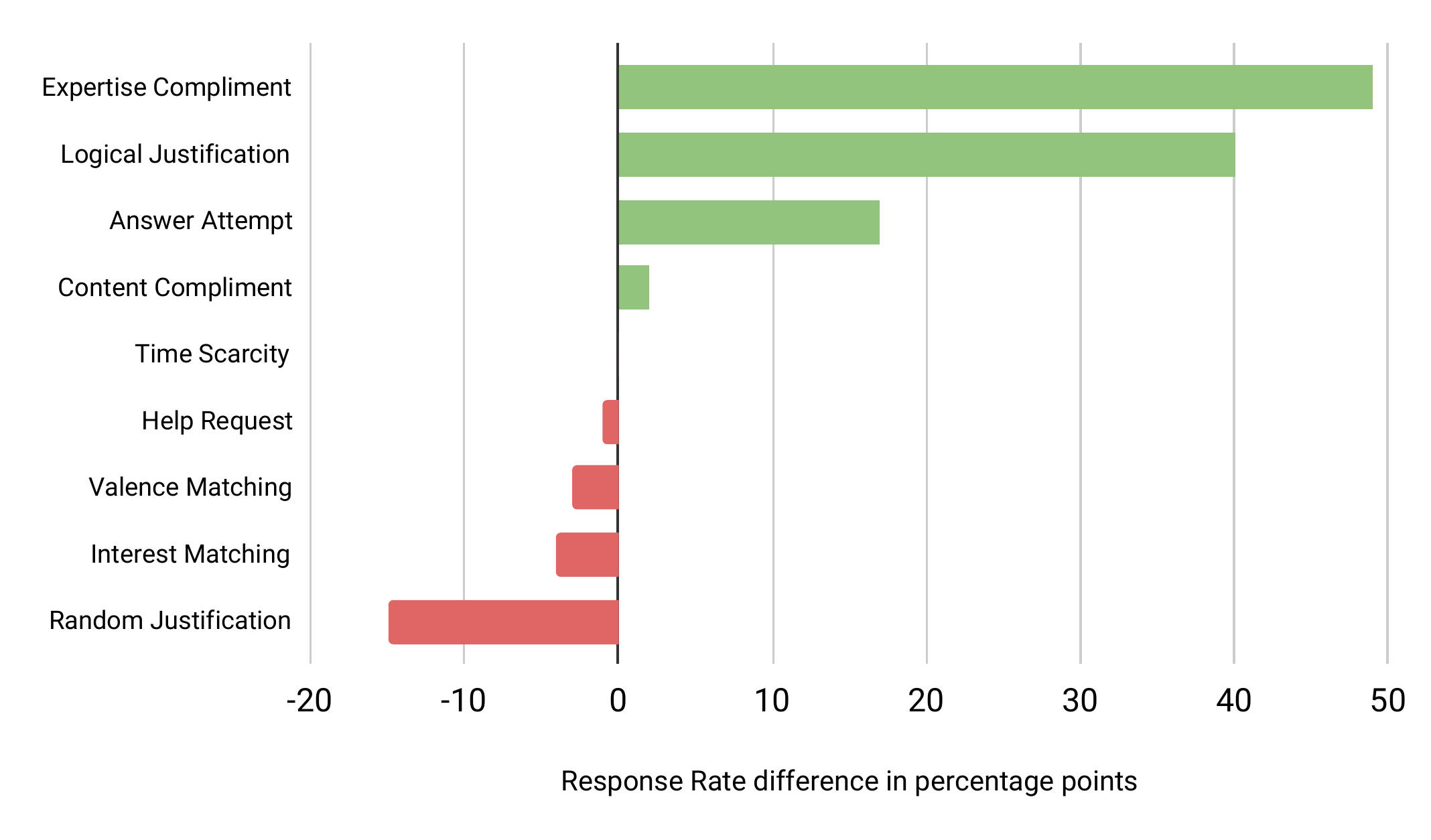}
\caption{Difference between response rate of the agent and humans for each social strategy. Green indicates the agent is better than people and red indicates the opposite.}
\label{fig:AgentVsHuman}
\end{figure}

\runinhead{Strategies surveyed.} We use all nine strategies described earlier and add a baseline and an oracle strategy. The baseline case posts the raw question with no augmentation. The oracle method asks AMT workers to modify the question without any restrictions to maximize the chances of receiving the answer. They don't have to follow any of our outlined social strategies.

\runinhead{Dataset of online interactions.} To study the impact of using social strategies, we collect a dataset of $10$k posts for each of the $4$ ML social strategies, resulting in a dataset of $40$k questions with augmentations. The $5$ rule strategies don't require any training data. Once trained, we post $100$ questions per strategy to Instagram, resulting in $1100$ total posts. To further study the scalability and transfer of strategies learned in one domain and applied to another, we train augmentation models using data from a ``source'' domain and test its effect on posts from ``target'' domains. For example, we train models using data collected from the $\#$travel source domain and test on the rest as target domains.

To train the selection model, we gather $10$k posts from Instagram and generate augmentations with each of the social strategies. Each post, with all the augmentated questions, is sent to AMT workers, who are asked to pick the strategies that would be appropriate to use. We choose to train the selection model using AMT instead of Instagram as it allows us to quickly collect large amounts of training data and negate the impact of other confounds. Each AMT task included $10$ social media posts. One out of the ten posts contained an attention checker in the question to verify that the workers were actually reading the questions. Workers were compensated at a rate of $\$12$ per hour.

\subsubruninhead{Augmenting questions with social strategies}
Our goal in the first set of experiments is to study the effect of using social strategies to augment questions.

\runinhead{Informative responses.}
Before we inspect the effects of social strategies, we first report the quality of responses from Instagram users.
We manually annotate all our responses and find that $93.01\%$ of questions are both relevant as well as answerable. Out of the relevant questions, $95.52\%$ of responses were informative, i.e.~the responses contained the correct answer to the question. Figure~\ref{fig:ESResponse2} visualizes a set of example responses for different posts with different social strategies in the travel domain. While all social strategies outperformed the baseline in receiving responses, the quality of the responses differed across strategies.

\runinhead{Effect of social strategies.}
Table~\ref{tab:crossdomain} reports the informative response rate across all the social strategies. We find that, compared to the baseline case, where no strategy is used, rule-based strategies improve participation by $14.78$ percent points. An unpaired t-test confirms that participation increases by designing appropriate rule-based social strategies ($t(900)=3.05$, $p<0.01$). When social strategy data is collected and used to train ML strategies, performance increases by $42.36$ percent points and $27.58$ percent points when compared against un-augmented ($t(900)=8.17$, $p<0.001$) and rule-based strategies ($t(900)=8.96$, $p<0.001$) and confirmed by unpaired t-tests. Overall, we find that \texttt{expertise compliment} and \texttt{logical justification} performed strongly in shopping domain, but weakly in animals and food domains.

\begin{figure}[t]
\centering
\includegraphics[width=\linewidth]{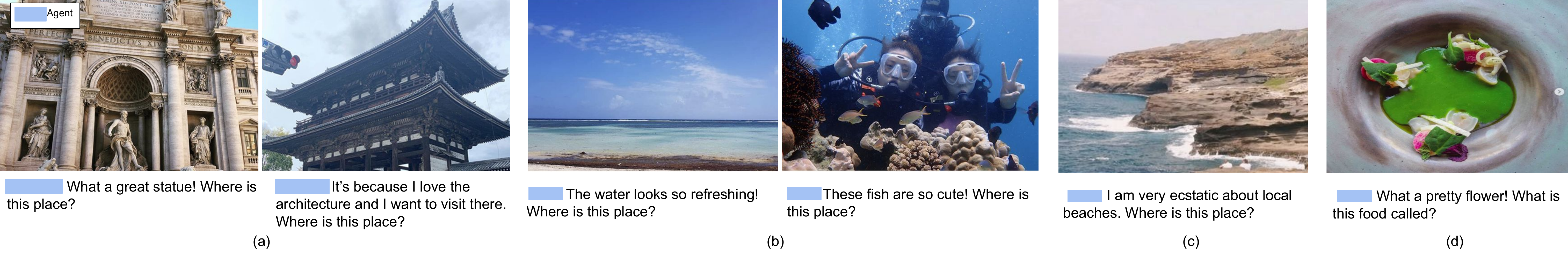}
\caption{Example strategy selection and augmentations in the travel domain. (a) Our system learns to focus on different aspects of the image. (b) The system is able to discern between very similar images and understand that the same objects can have different connotations. (c, d) Example failure case when objects were misclassified.}
\label{fig:genexamples}
\end{figure}

To test the scalability of our strategies across image domains, we train models on a source domain and deploy them on a target domain. We find that \texttt{expertise compliment} drops in performance while \texttt{interest matching} improves. The drop implies that machine learning models that heavily depend on example data points used in training process are not robust in new domains. Therefore, while machine learning strategies are the most effective, they require strategy data collected for the domain in which they are deployed. The drop in performance, however, still results in improvements in response rate, demonstrating that machine learning strategies scale across domains but their impact reduces as the distribution of image content increases from the source domain. The increase in performance of \texttt{interest matching} indicates that different domains might have different dominating social strategies, i.e.~no single dominant strategy exists across all domains and that a selection component is necessary.

\runinhead{Agent versus human augmentations.} 
We compare the augmentations generated by our agent against those created by crowdworkers. We report the difference in response rate between the agent and the human augmentations across the different strategies in Figure~\ref{fig:AgentVsHuman}.
A two-way ANOVA finds that the strategy used has a significant effect on the response rate ($F(8,900) = 12.99$, $p<0.001$) but the poster has no significant effect on the response rate ($F(1,900)=1.82$, $p=0.17$). The ANOVA also found a significant interaction effect between the strategy and the poster on response rate ($F(1,900)=2.09$, $p=0.03$). A posthoc Tukey test indicates that the agent using the machine learning strategies is significantly increases response rate than the agent using rule-based  ($p < 0.05$) or humans using rule-based strategies ($p < 0.05$). This demonstrates that a machine learning model that has witnessed examples of social strategies can outperform rule-based systems. However, there is no significant difference between the agent using machine learning strategies versus humans using the same social strategies.


\subsubruninhead{Learning to select a social strategy}
In our previous experiment we established that different domains have different strategies that perform best. Now, we evaluate how well our selection component performs at selecting the most effective strategy. Specifically, we test how well our selection model performs (1) against a random strategy, (2) against the most effective strategy (\texttt{expertise compliment}) from the previous experiment, and (3) against the oracle strategy generated by crowdworkers. Recall that the oracle strategy does not constrain workers to use any particular strategy. 

Since this test needs be able to test multiple strategies on the same post, we perform our evaluation on AMT. Workers are shown two strategies for a given post and asked to choose which strategy is most likely to receive a response. We perform pairwise comparisons between our selection model against a random strategy across $11$k posts, against \texttt{expertise compliment} across $549$ posts and against open-ended human questions across $689$ posts.

\runinhead{Effect of selection.} A binomial test indicates that our selection method was chosen $54.12\%$ more often than a random strategy $B(N=11,844, p<0.001)$. It was chosen $58.28\%$ more often than \texttt{expertise compliment} $B(N=549, p<0.001)$. And finally, it was chosen $75.61\%$ more often than the oracle human generated questions $B(N=689, p<0.001)$. We conclude that our selection model outperforms all existing baselines.

\runinhead{Qualitative analysis.}
Figure~\ref{fig:genexamples}(a) shows that the agent can choose to focus on different aspects of the image even when the subject of the image is roughly the same: old traditional buildings. In one, the agent compliments the statue, which is the most salient feature of the old European building shown in the image. In the other, it shows appreciation for the overall architecture of the old Asian building, which does not have a single defining feature like a statue.

Figure \ref{fig:genexamples}(b) shows two images that are both contain water and has similar color composition. In one, the agent compliments the water seen on the beach as refreshing and in the other, the fish seen underwater as cute. Referring to a fish in a beach photo would have been incorrect as would have been describing water as refreshing in an underwater photo. 

Though social strategies are useful, they can also lead to new errors. Figure~\ref{fig:genexamples}(c, d) showcases an example questions where the agent fails to recognize mountains and food and generates phrases referring to beaches and flowers.

\subsection{Discussion}

\runinhead{Intended use.} This work demonstrates that it is possible to train an AI agent to use social strategies that are found in human-to-human interaction contexts to increase the likelihood of a human crowdsourcing respondent. 
Such responses suggest a future in which supervised ML models can be trained on authentic online data that are provided by willing helpers than from paid workers. We expect that such strategies can lead to adaptive ML systems that can learn during their deployment, by asking their users whenever they are uncertain about their environment. Unlike existing paid crowdsourcing techniques that grow linearly in cost as the number of annotations increases, our method is a fixed cost solution where social strategies need to be collected for a specific domain and then deployed to encourage volunteers.

\runinhead{Negative usage.} It is also important that we pause to note the potential negative implications of computing research, and how they can be addressed. The psychology techniques that our work relies on have been used in negotiations and marketing campaigns for decades. Automating such techniques can also lead to influencing emotions or behavior at a magnitude greater than single human-human interaction~\cite{kramer2014experimental,ferrara2016rise}. When using natural language techniques, we advocate that agents continue to self-identify as bots for this reason. There is a need for online communities to establish a standard acceptable use of such techniques and how the contributors should be informed about the intentions behind an agent's request.

\runinhead{Limitations and future work.} Our social strategies are inspired by social psychology research. Ours are by no means an exhaustive list of possible strategies. Future research could follow a more ``bottom-up'' approach of directly learning to emulate strategies by observing human-human interactions. Currently, our requests involve exactly one dialogue turn, and we do not yet explore multi-turn conversations. This can be important: for example, the answer attempt strategy may be more effective at getting an answer now, but might also decrease the probability that the contributor will want to continue cooperating in the long term. Future work can explore how to guide conversations to enable more complex labeling schemes.

\runinhead{Conclusion}
Our work: (1)~identifies social strategies that can be repurporsed to improve crowdsourcing requests for visual question answering, (2)~trains and deploys machine learning and rule-based models that deploy these strategies to increase crowdsourcing participation, and (3)~demonstrates that these models significantly improve participation on Instagram, that no single strategy is optimal, and that a selection model can chooses the appropriate strategy.

\section{Model evaluation using human perception}
\label{sec:4}
Generating realistic images is regarded as a focal task for measuring the progress of generative models. Automated metrics are either heuristic approximations~\cite{rossler2019faceforensics++,salimans2016improved,denton2015deep,karras2018style,brock2018large,radford2015unsupervised} or intractable density estimations, examined to be inaccurate on high dimensional problems~\cite{hinton2002training,bishop2006pattern,theis2015note}. Human evaluations, such as those given on Amazon Mechanical Turk~\cite{rossler2019faceforensics++,denton2015deep}, remain ad-hoc because ``results change drastically''~\cite{salimans2016improved} based on details of the task design~\cite{liu2016effective,le2010ensuring,kittur2008crowdsourcing}. With both noisy automated and noisy human benchmarks, measuring progress over time has become akin to hill-climbing on noise. Even widely used metrics, such as Inception Score~\cite{salimans2016improved} and Fr\'echet Inception Distance~\cite{heusel2017gans}, have been discredited for their application to non-ImageNet datasets~\cite{barratt2018note,rosca2017variational,borji2018pros,ravuri2018learning}. Thus, to monitor progress, generative models need a systematic gold standard benchmark. In this paper, we introduce a gold standard benchmark for realistic generation, demonstrating its effectiveness across four datasets, six models, and two sampling techniques, and using it to assess the progress of generative models over time.

\begin{figure}[!htb]
    \center{\includegraphics[width=.8\textwidth]{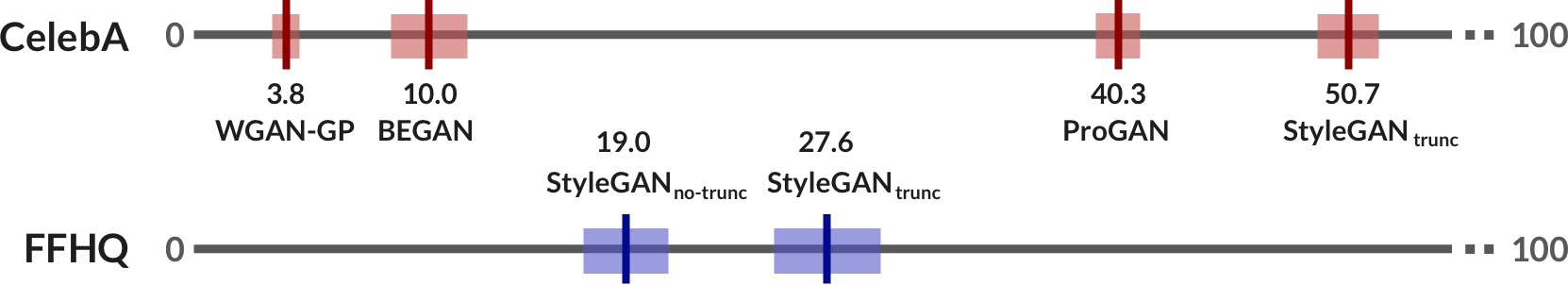}}
    \caption{\label{fig:pull} Our human evaluation metric, \system, consistently distinguishes models from each other: here, we compare different generative models performance on FFHQ.  A score of $50\%$ represents indistinguishable results from real, while a score above $50\%$ represents hyper-realism.}
\end{figure}

Realizing the constraints of available automated metrics, many generative modeling tasks resort to human evaluation and visual inspection~\cite{rossler2019faceforensics++,salimans2016improved, denton2015deep}. These human measures are (1)~ad-hoc, each executed in idiosyncrasy without proof of reliability or grounding to theory, and (2)~high variance in their estimates~\cite{salimans2016improved, denton2015deep, olsson2018skill}. These characteristics combine to a lack of reliability, and downstream, (3)~a lack of clear separability between models. Theoretically, given sufficiently large sample sizes of human evaluators and model outputs, the law of large numbers would smooth out the variance and reach eventual convergence; but this would occur at (4)~a high cost and a long delay.

We present \system (\systemdesc) to address these criteria in turn. \system: (1)~measures the perceptual realism of generative model outputs via a \textbf{grounded} method inspired by psychophysics methods in perceptual psychology, (2)~is a \textbf{reliable} and consistent estimator, (3)~is statistically \textbf{separable} to enable a comparative ranking, and (4)~ensures a cost and time \textbf{efficient} method through modern crowdsourcing techniques such as training and aggregation. We present two methods of evaluation. The first, called \systemstair, is inspired directly by the psychophysics literature~\cite{klein2001measuring,cornsweet1962staircrase}, and displays images using adaptive time constraints to determine the time-limited perceptual threshold a person needs to distinguish real from fake. The \systemstair score is understood as the minimum time, in milliseconds, that a person needs to see the model's output before they can distinguish it as real or fake. For example, a score of $500$ms on \systemstair indicates that humans can distinguish model outputs from real images at $500$ms exposure times or longer, but not under $500$ms.
The second method, called \systeminf, is derived from the first to make it simpler, faster, and cheaper while maintaining reliability. It is interpretable as the rate at which people mistake fake images and real images, given unlimited time to make their decisions. A score of $50\%$ on \systeminf means that people differentiate generated results from real data at chance rate, while a score above $50\%$ represents hyper-realism in which generated images appear more real than real images.

We run two large-scale experiments. First, we demonstrate \system's performance on unconditional human face generation using four popular generative adversarial networks (GANs)~\cite{gulrajani2017improved,berthelot2017began,karras2017progressive,karras2018style} across \taskceleba~\cite{liu2015faceattributes}. We also evaluate two newer GANs~\cite{miyato2018spectral,brock2018large} on \taskffhq~\cite{karras2018style}. \system indicates that GANs have clear, measurable perceptual differences between them; this ranking is identical in both \systemstair and \systeminf. The best performing model, StyleGAN trained on FFHQ and sampled with the truncation trick, only performs at $27.6\%$ \systeminf, suggesting substantial opportunity for improvement. We can reliably reproduce these results with $95\%$ confidence intervals using $30$ human evaluators at $\$60$ in a task that takes $10$ minutes.

Second, we demonstrate the performance of \systeminf beyond faces on conditional generation of five object classes in ImageNet~\cite{deng2009imagenet} and unconditional generation of \taskcifar~\cite{krizhevsky2009learning}. Early GANs such as BEGAN are not separable in \systeminf when generating \taskcifar: none of them produce convincing results to humans, verifying that this is a harder task than face generation. The newer StyleGAN shows separable improvement, indicating progress over the previous models. With \taskimagenet, GANs have improved on classes considered ``easier'' to generate (e.g.,~lemons), but resulted in consistently low scores across all models for harder classes (e.g.,~French horns).

\system is a rapid solution for researchers to measure their generative models, requiring just a single click to produce reliable scores and measure progress. We deploy \system at \website, where researchers can upload a model and retrieve a \system score. Future work will extend \system to additional generative tasks such as text, music, and video generation.


\subsection{\system: A benchmark for \systemdesc}
\label{sec:approach}

\system displays a series of images one by one to crowdsourced evaluators on Amazon Mechanical Turk and asks the evaluators to assess whether each image is real or fake. Half of the images are real images, drawn from the model's training set (e.g.,~FFHQ, CelebA, ImageNet, or CIFAR-10). The other half are drawn from the model's output. We use modern crowdsourcing training and quality control techniques~\cite{mitra2015comparing} to ensure high-quality labels. Model creators can choose to perform two different evaluations: \systemstair, which gathers time-limited perceptual thresholds to measure the psychometric function and report the minimum time people need to make accurate classifications, and \systeminf, a simplified approach which assesses people's error rate under no time constraint.

\begin{figure}[h]
\centering
\sidecaption
\includegraphics[width=.55\textwidth]{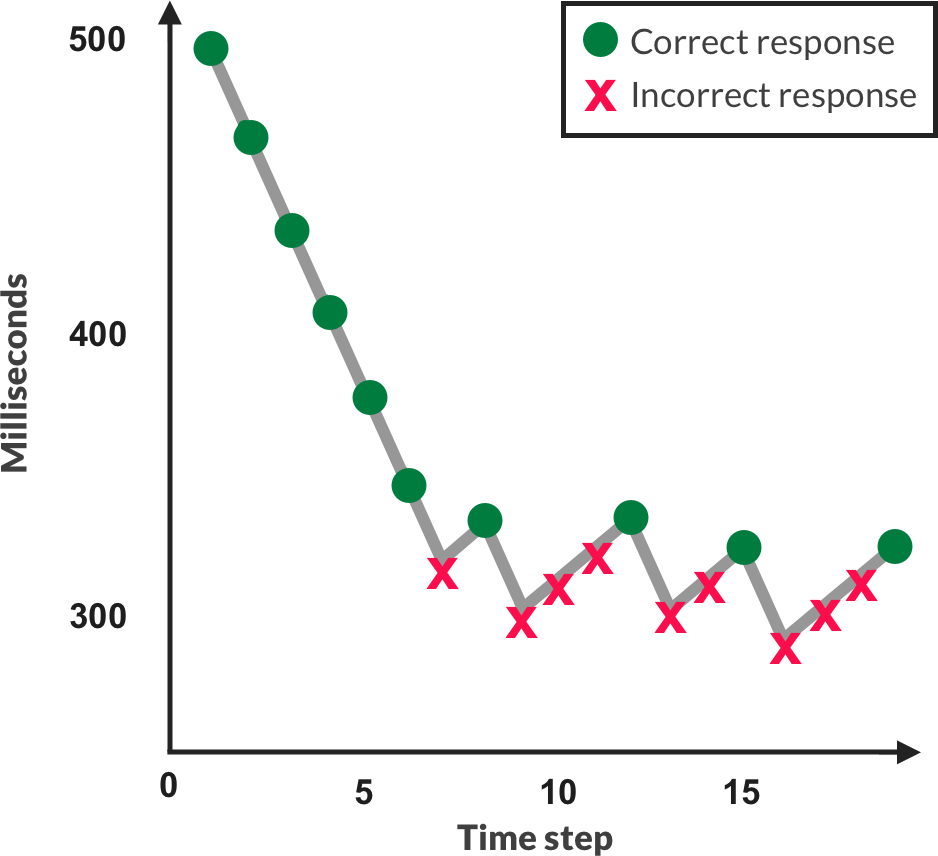}
\caption{The adaptive staircase method shows images to evaluators at different time exposures, decreasing when correct and increasing when incorrect. The modal exposure measures their perceptual threshold.}
\label{fig:staircase-generic} 
\end{figure}

\subsubruninhead{\systemstair: Perceptual fidelity grounded in psychophysics}
Our first method, \systemstair, measures time-limited perceptual thresholds. It is rooted in psychophysics literature, a field devoted to the study of how humans perceive stimuli, to evaluate human time thresholds upon perceiving an image. Our evaluation protocol follows the procedure known as the \textit{adaptive staircase method} (Figure~\ref{fig:staircase-generic})~\cite{cornsweet1962staircrase}. An image is flashed for a limited length of time, after which the evaluator is asked to judge whether it is real or fake. If the evaluator consistently answers correctly, the staircase descends and flashes the next image with less time. If the evaluator is incorrect, the staircase ascends and provides more time. 

\begin{figure}[tb]
    \center{\includegraphics[width=.95\textwidth]{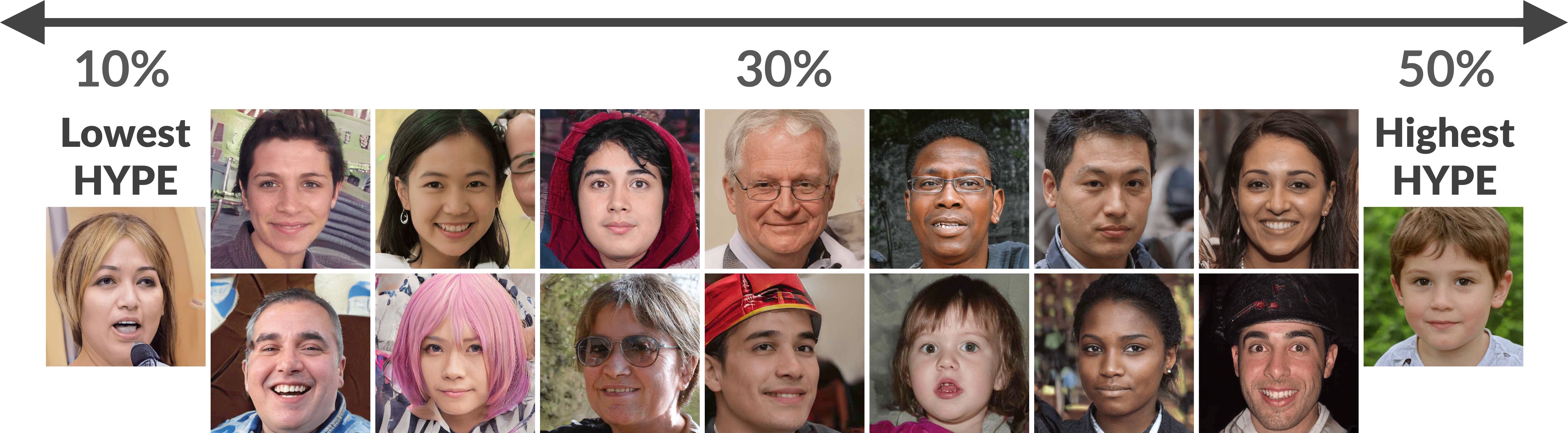}}
    \caption{\label{fig:examples} Example images sampled with the truncation trick from StyleGAN trained on FFHQ. Images on the right exhibit the highest \systeminf scores, the highest human perceptual fidelity.}
\end{figure}

This process requires sufficient iterations to converge to the evaluator's perceptual threshold: the shortest exposure time at which they can maintain effective performance~\cite{cornsweet1962staircrase,greene2009briefest,fei2007we}. The process produces what is known as the \textit{psychometric function}~\cite{wichmann2001psychometric}, the relationship of timed stimulus exposure to accuracy. For example, for an easily distinguishable set of generated images, a human evaluator would immediately drop to the lowest millisecond exposure.

\systemstair displays three blocks of staircases for each evaluator. An image evaluation begins with a \mbox{3-2-1} countdown clock, each number displaying for $500$ms~\cite{krishna2016embracing}. The sampled image is then displayed for the current exposure time. Immediately after each image, four perceptual mask images are rapidly displayed for $30$ms each. These noise masks are distorted to prevent retinal afterimages and further sensory processing after the image disappears~\cite{greene2009briefest}. We generate masks using an existing texture-synthesis algorithm~\cite{portilla2000parametric}. Upon each submission, \systemstair reveals to the evaluator whether they were correct.

Image exposures are in the range [$100$ms, $1000$ms], derived from the perception literature~\cite{fraisse1984perception}. All blocks begin at $500$ms and last for $150$ images ($50$\% generated, $50$\% real), values empirically tuned from prior work~\cite{cornsweet1962staircrase,dakin2009psychophysical}. Exposure times are raised at $10$ms increments and reduced at $30$ms decrements, following the $3$-up/$1$-down adaptive staircase approach, which theoretically leads to a $75\%$ accuracy threshold that approximates the human perceptual threshold~\cite{levitt1971transformed,greene2009briefest,cornsweet1962staircrase}.

Every evaluator completes multiple staircases, called \textit{blocks}, on different sets of images. As a result, we observe multiple measures for the model. We employ three blocks, to balance quality estimates against evaluators' fatigue~\cite{krueger1989sustained,rzeszotarski2013inserting,hata2017glimpse}. We average the modal exposure times across blocks to calculate a final value for each evaluator. Higher scores indicate a better model, whose outputs take longer time exposures to discern from real.

\subsubruninhead{\systeminf: Cost-effective approximation}
Building on the previous method, we introduce \systeminf: a simpler, faster, and cheaper method after ablating \systemstair to optimize for speed, cost, and ease of interpretation. \systeminf shifts from a measure of perceptual time to a measure of human deception rate, given infinite evaluation time. The \systeminf score gauges total error on a task of 50 fake and 50 real images \footnote{We explicitly reveal this ratio to evaluators. Amazon Mechanical Turk forums would enable evaluators to discuss and learn about this distribution over time, thus altering how different evaluators would approach the task. By making this ratio explicit, evaluators would have the same prior entering the task.}, enabling the measure to capture errors on both fake and real images, and effects of hyperrealistic generation when fake images look even more realistic than real images \footnote{Hyper-realism is relative to the real dataset on which a model is trained. Some datasets already look less realistic because of lower resolution and/or lower diversity of images.}. \systeminf requires fewer images than \systemstair to find a stable value, empirically producing a $6$x reduction in time and cost ($10$ minutes per evaluator instead of $60$ minutes, at the same rate of $\$12$ per hour). Higher scores are again better: $10\%$ \systeminf indicates that only $10\%$ of images deceive people, whereas $50\%$ indicates that people are mistaking real and fake images at chance, rendering fake images indistinguishable from real. Scores above $50\%$ suggest hyperrealistic images, as evaluators mistake images at a rate greater than chance.

\systeminf shows each evaluator a total of $100$ images: $50$ real and $50$ fake. We calculate the proportion of images that were judged incorrectly, and aggregate the judgments over the $n$ evaluators on $k$ images to produce the final score for a given model. 

\subsection{Consistent and reliable design}
To ensure that our reported scores are consistent and reliable, we need to sample sufficiently from the model as well as hire, qualify, and appropriately pay enough evaluators.

\runinhead{Sampling sufficient model outputs.}
The selection of $K$ images to evaluate from a particular model is a critical component of a fair and useful evaluation. We must sample a large enough number of images that fully capture a model's generative diversity, yet balance that against tractable costs in the evaluation. We follow existing work on evaluating generative output by sampling $K=5000$ generated images from each model~\cite{salimans2016improved, miyato2018spectral, warde2016improving} and $K=5000$ real images from the training set. From these samples, we randomly select images to give to each evaluator.

\runinhead{Quality of evaluators.} To obtain a high-quality pool of evaluators, each is required to pass a qualification task. Such a pre-task filtering approach, sometimes referred to as a person-oriented strategy, is known to outperform process-oriented strategies that perform post-task data filtering or processing~\cite{mitra2015comparing}. Our qualification task displays $100$ images ($50$ real and $50$ fake) with no time limits. Evaluators must correctly classify $65\%$ of both real and fake images. This threshold should be treated as a hyperparameter and may change depending upon the GANs used in the tutorial and the desired discernment ability of the chosen evaluators. We choose $65\%$ based on the cumulative binomial probability of 65 binary choice answers out of 100 total answers: there is only a one in one-thousand chance that an evaluator will qualify by random guessing. Unlike in the task itself, fake qualification images are drawn equally from multiple different GANs to ensure an equitable qualification across all GANs. The qualification is designed to be taken occasionally, such that a pool of evaluators can assess new models on demand.

\runinhead{Payment.} Evaluators are paid a base rate of $\$1$ for working on the qualification task. To incentivize evaluators to remained engaged throughout the task, all further pay after the qualification comes from a bonus of $\$0.02$ per correctly labeled image, typically totaling a wage of $\$12$/hr.

\subsection{Experimental setup}
\label{sec:experiments}

\runinhead{Datasets.} We evaluate on four datasets. (1) \taskceleba~\cite{liu2015faceattributes} is popular dataset for unconditional image generation with $202$k images of human faces, which we align and crop to be $64 \times 64$ px. (2) \taskffhq~\cite{karras2018style} is a newer face dataset with $70$k images of size $1024\times1024$ px. (3) \taskcifar consists of $60$k images, sized $32\times32$ px, across $10$ classes. (4) \taskimagenet is a subset of $5$ classes with $6.5$k images at $128\times128$ px from the ImageNet dataset~\cite{deng2009imagenet}, which have been previously identified as easy (lemon, Samoyed, library) and hard (baseball player, French horn)~\cite{brock2018large}.

\runinhead{Architectures.} We evaluate on four state-of-the-art models trained on \taskceleba and \taskcifar: StyleGAN~\cite{karras2018style}, ProGAN~\cite{karras2017progressive}, BEGAN~\cite{berthelot2017began}, and WGAN-GP~\cite{gulrajani2017improved}. We also evaluate on two models, SN-GAN~\cite{miyato2018spectral} and BigGAN~\cite{brock2018large} trained on ImageNet, sampling conditionally on each class in \taskimagenet. We sample BigGAN with ($\sigma=0.5$~\cite{brock2018large}) and without the truncation trick.

We also evaluate on StyleGAN~\cite{karras2018style} trained on \taskffhq with ($\psi=0.7$~\cite{karras2018style}) and without truncation trick sampling. For parity on our best models across datasets, StyleGAN instances trained on \taskceleba and \taskcifar are also sampled with the truncation trick.


We sample noise vectors from the $d$-dimensional spherical Gaussian noise prior $z\in \mathbb{R}^{d} \sim \mathcal{N}(0,I)$ during training and test times. We specifically opted to use the same standard noise prior for comparison, yet are aware of other priors that optimize for FID and IS scores~\cite{brock2018large}. We select training hyperparameters published in the corresponding papers for each model.

\runinhead{Evaluator recruitment.} We recruit $930$ evaluators from Amazon Mechanical Turk, or 30 for each run of \system. To maintain a between-subjects study in this evaluation, we recruit independent evaluators across tasks and methods. 

\runinhead{Metrics.} For \systemstair, we report the modal perceptual threshold in milliseconds. For \systeminf, we report the error rate as a percentage of images, as well as the breakdown of this rate on real and fake images separately. To show that our results for each model are separable, we report a one-way ANOVA with Tukey pairwise post-hoc tests to compare all models. 

Reliability is a critical component of \system, as a benchmark is not useful if a researcher receives a different score when rerunning it. We use bootstrapping~\cite{felsenstein1985confidence}, repeated resampling from the empirical label distribution, to measure variation in scores across multiple samples with replacement from a set of labels. We report $95\%$ bootstrapped confidence intervals (CIs), along with standard deviation of the bootstrap sample distribution, by randomly sampling $30$ evaluators with replacement from the original set of evaluators across $10,000$ iterations.

\begin{table}[h]
\centering\begin{tabular}[t]{clccc}
Rank & GAN & \textbf{\systemstair (ms)} & Std. &  95\% CI       \\
\hline
1 & \trunc & 363.2     & 32.1         & 300.0 -- 424.3 \\
2 & \notrunc   & 240.7    & 29.9          & 184.7 -- 302.7 \\
\hline
\end{tabular}
\caption{\systemstair on \trunc and \notrunc trained on \taskffhq.}
\label{table:stair_ffhq}
\end{table}

\runinhead{Experiment 1:} We run two large-scale experiments to validate \system. The first one focuses on the controlled evaluation and comparison of \systemstair against \systeminf on established human face datasets. We recorded responses totaling ($4$ \taskceleba $+$ $2$ \taskffhq) models $\times$ $30$ evaluators $\times$ $550$ responses = $99$k total responses for our \systemstair evaluation and ($4$ \taskceleba $+$ $2$ \taskffhq) models $\times$ $30$ evaluators $\times$ $100$ responses = $18$k, for our \systeminf evaluation.

\runinhead{Experiment 2:} The second experiment evaluates \systeminf on general image datasets. We recorded ($4$ \taskcifar $+$ $3$ \taskimagenet) models $\times$ $30$ evaluators $\times$ $100$ responses = $57$k total responses.


\subsection{Experiment 1: \systemstair and \systeminf on human faces}
\label{sec:results}

We report results on \systemstair and demonstrate that the results of \systeminf approximates those from \systemstair at a fraction of the cost and time.

\subsubruninhead{\systemstair}

\taskceleba We find that \trunc resulted in the highest \systemstair score (modal exposure time), at a mean of $439.3$ms, indicating that evaluators required nearly a half-second of exposure to accurately classify \trunc 
images (Table~\ref{table:stair_ffhq}). 
\trunc is followed by ProGAN at $363.7$ms, a $17\%$ drop in time. BEGAN and WGAN-GP are both easily identifiable as fake, tied in last place around the minimum available exposure time of $100$ms. 
Both BEGAN and WGAN-GP exhibit a bottoming out effect --- reaching the minimum time exposure of $100$ms quickly and consistently.

To demonstrate separability between models we report results from a one-way analysis of variance (ANOVA) test, where each model's input is the list of modes from each model's $30$ evaluators. The ANOVA results confirm that there is a statistically significant omnibus difference ($F(3, 29) = 83.5, p < 0.0001$). Pairwise post-hoc analysis using Tukey tests confirms that all pairs of models are separable (all $p<0.05$) except BEGAN and WGAN-GP ($n.s.$).

\runinhead{\taskffhq}. We find that \trunc resulted in a higher exposure time than \notrunc, at $363.2$ms and $240.7$ms, respectively (Table~\ref{table:stair_ffhq}). While the $95\%$ confidence intervals that represent a very conservative overlap of $2.7$ms, an unpaired t-test confirms that the difference between the two models is significant ($t(58)=2.3, p = 0.02$).

\subsubruninhead{\systeminf}
\runinhead{\taskceleba}. Table \ref{table:inf_celeba} reports results for \systeminf on \taskceleba. We find that \trunc resulted in the highest \systeminf score, fooling evaluators $50.7\%$ of the time. \trunc is followed by ProGAN at $40.3\%$, BEGAN at $10.0\%$, and WGAN-GP at $3.8\%$. No confidence intervals are overlapping and an ANOVA test is significant ($F(3, 29) = 404.4, p < 0.001$). Pairwise post-hoc Tukey tests show that all pairs of models are separable (all $p<0.05$). Notably, \systeminf results in separable results for BEGAN and WGAN-GP, unlike in \systemstair where they were not separable due to a bottoming-out effect.

\begin{table}[h]
\centering
\resizebox{\textwidth}{!}{\begin{tabular}[tb]{clcccccccc}
Rank & GAN & \textbf{\systeminf (\%)} & Fakes Error & Reals Error & Std. & 95\% CI  & KID & FID & Precision\\
\hline
1 & \trunc & 50.7\%   & 62.2\% & 39.3\% &  1.3   & 48.2 -- 53.1 & 0.005 & 131.7 & 0.982 \\
2 &  ProGAN   & 40.3\%  &  46.2\% & 34.4\% &  0.9          & 38.5 -- 42.0 & 0.001 & 2.5 &  0.990 \\
3 & BEGAN    & 10.0\%   &  6.2\% & 13.8\% &  1.6         &  7.2 -- 13.3   & 0.056 & 67.7 & 0.326       \\
4 & WGAN-GP  & 3.8\%    &  1.7\% &  5.9\%    &  0.6    & 3.2 -- 5.7   & 0.046 & 43.6 & 0.654       \\
\hline
\end{tabular}}
\caption{\systeminf on four GANs trained on \taskceleba. Counterintuitively, real errors increase with the errors on fake images, because evaluators become more confused and distinguishing factors between the two distributions become harder to discern.}
\label{table:inf_celeba}
\end{table}

\runinhead{\taskffhq}. We observe a consistently separable difference between \trunc and \notrunc and clear delineations between models (Table~\ref{table:inf_ffhq}). \systeminf ranks \trunc ($27.6\%$) above \notrunc ($19.0\%$) with no overlapping CIs. Separability is confirmed by an unpaired t-test ($t(58)=8.3, p < 0.001$).

\begin{table}[h]
\centering
\resizebox{\textwidth}{!}{\begin{tabular}[t]{clcccccccc}
Rank & GAN & \textbf{\systeminf (\%)} & Fakes Error & Reals Error & Std. & 95\% CI & KID & FID & Precision     \\
\hline
1 & \trunc & 27.6\%   & 28.4\% & 26.8\%  &2.4    & 22.9 -- 32.4 & 0.007 & 13.8 & 0.976\\
2 & \notrunc   & 19.0\%   &18.5\% & 19.5\%  &1.8         & 15.5 -- 22.4 & 0.001 & 4.4 & 0.983\\
\hline
\end{tabular}}
\caption{\systeminf on \trunc and \notrunc trained on \taskffhq. Evaluators were deceived most often by \trunc. Similar to \taskceleba, fake errors and real errors track each other as the line between real and fake distributions blurs.}
\label{table:inf_ffhq}
\end{table}

\subsubruninhead{Cost tradeoffs with accuracy and time}\label{sec:tradeoffs}

\begin{figure}[h]
\centering
\includegraphics[width=.65\textwidth]{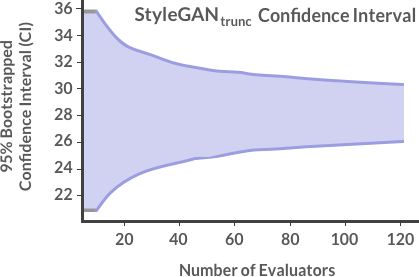}
\caption{Effect of more evaluators on CI.}
\label{fig:worker-experiment}
\end{figure}

One of \system's goals is to be cost and time efficient. When running \system, there is an inherent tradeoff between accuracy and time, as well as between accuracy and cost. This is driven by the law of large numbers: recruiting additional evaluators in a crowdsourcing task often produces more consistent results, but at a higher cost (as each evaluator is paid for their work) and a longer amount of time until completion (as more evaluators must be recruited and they must complete their work).

To manage this tradeoff, we run an experiment with \systeminf on \trunc. We completed an additional evaluation with $60$ evaluators, and compute $95\%$ bootstrapped confidence intervals, choosing from $10$ to $120$ evaluators (Figure~\ref{fig:worker-experiment}). We see that the CI begins to converge around $30$ evaluators, our recommended number of evaluators to recruit.

At $30$ evaluators, the cost of running \systemstair on one model was approximately $\$360$, while the cost of running \systeminf on the same model was approximately $\$60$. Payment per evaluator for both tasks was approximately $\$12$/hr. Evaluators spent an average of one hour each on a \systemstair task and $10$ minutes each on a \systeminf task. Thus, \systeminf achieves its goals of being significantly cheaper to run, while maintaining consistency.

\subsubruninhead{Comparison to automated metrics}
As FID \cite{heusel2017gans} is one of the most frequently used evaluation methods for unconditional image generation, it is imperative to compare \system against FID on the same models. We also compare to two newer automated metrics: KID~\cite{binkowski2018demystifying}, an unbiased estimator independent of sample size, and $F_{1/8}$ (precision)~\cite{sajjadi2018assessing}, which captures fidelity independently.
We show through Spearman rank-order correlation coefficients that \system scores are not correlated with FID ($\rho=-0.029, p=0.96$), where a Spearman correlation of $-1.0$ is ideal because lower FID and higher \system scores indicate stronger models. We therefore find that FID is not highly correlated with human judgment. Meanwhile, \systemstair and \systeminf exhibit strong correlation ($\rho=1.0, p=0.0$), where $1.0$ is ideal because they are directly related. We calculate FID across the standard protocol of $50$K generated and $50$K real images for both \taskceleba and \taskffhq, reproducing scores for \notrunc. KID ($\rho=-0.609, p=0.20$) and precision ($\rho=0.657, p=0.16$) both show a statistically insignificant but medium level of correlation with humans.

\subsubruninhead{\systeminf during model training}
\system can also be used to evaluate progress during model training. We find that \systeminf scores increased as StyleGAN training progressed from $29.5\%$ at $4$k epochs, to $45.9\%$ at $9$k epochs, to $50.3\%$ at $25$k epochs ($F(2, 29) = 63.3, p < 0.001$). 

\subsection{Experiment 2: \systeminf beyond faces}
We now turn to another popular image generation task: objects. As Experiment 1 showed \systeminf to be an efficient and cost effective variant of \systemstair, here we focus exclusively on \systeminf.

\subsubruninhead{\taskimagenet}

We evaluate conditional image generation on five ImageNet classes (Table~\ref{table:imagenet-5}). We also report FID~\cite{heusel2017gans}, KID~\cite{binkowski2018demystifying}, and $F_{1/8}$ (precision)~\cite{sajjadi2018assessing} scores. To evaluate the relative effectiveness of the three GANs within each object class, we compute five one-way ANOVAs, one for each of the object classes. We find that the \systeminf scores are separable for images from three easy classes:
samoyeds (dogs) ($F(2, 29) = 15.0, p < 0.001$), lemons ($F(2, 29) = 4.2, p = 0.017$), and libraries ($F(2, 29) = 4.9, p = 0.009$). Pairwise Posthoc tests reveal that this difference is only significant between SN-GAN and the two BigGAN variants. We also observe that models have unequal strengths, e.g. SN-GAN is better suited to generating libraries than samoyeds.

\runinhead{Comparison to automated metrics}. Spearman rank-order correlation coefficients on all three GANs across all five classes show that there is a low to moderate correlation between the \systeminf scores and KID ($\rho=-0.377, p=0.02$), FID ($\rho=-0.282, p=0.01$), and negligible correlation with precision ($\rho=-0.067, p=0.81$). Some correlation for our \taskimagenet task is expected, as these metrics use pretrained ImageNet embeddings to measure differences between generated and real data. 

Interestingly, we find that this correlation depends upon the GAN: considering only SN-GAN, we find stronger coefficients for KID ($\rho=-0.500, p=0.39$), FID ($\rho=-0.300, p=0.62$), and precision ($\rho=-0.205, p=0.74$). When considering only BigGAN, we find far weaker coefficients for KID ($\rho=-0.151, p=0.68$), FID ($\rho=-0.067, p=.85$), and precision ($\rho=-0.164, p=0.65$). This illustrates an important flaw with these automatic metrics: their ability to correlate with humans depends upon the generative model that the metrics are evaluating on, varying by model and by task.

\begin{table}[h]
\centering
\resizebox{\textwidth}{!}{\begin{tabular}[tb]{cllcccccccc}
 & GAN & Class & \textbf{\systeminf (\%)} & Fakes Error & Reals Error & Std. & 95\% CI & KID & FID & Precision\\
\hline
\parbox[t]{2mm}{\multirow{3}{*}{\rotatebox[origin=c]{90}{Easy}}} & \BigGANtrunc &  Lemon &      18.4\% &      21.9\% &     14.9\% &  2.3 &  14.2--23.1 & 0.043 & 94.22 & 0.784\\
& \BigGANnotrunc & Lemon &      20.2\% &      22.2\% &     18.1\% &  2.2 &  16.0--24.8 & 0.036 & 87.54 & 0.774\\
& SN-GAN  &   Lemon &       12.0\% &      10.8\% &     13.3\% &  1.6 &   9.0--15.3 & 0.053 & 117.90 & 0.656\\
\hline
\parbox[t]{2mm}{\multirow{3}{*}{\rotatebox[origin=c]{90}{Easy}}}& \BigGANtrunc &  Samoyed &      19.9\% &      23.5\% &     16.2\% &  2.6 &  15.0--25.1 & 0.027 & 56.94 &0.794\\
& \BigGANnotrunc & Samoyed &      19.7\% &      23.2\% &     16.1\% &  2.2 &  15.5--24.1 & 0.014 & 46.14 & 0.906 \\
& SN-GAN  &   Samoyed &       5.8\% &       3.4\% &      8.2\% &  0.9 &    4.1--7.8 & 0.046 & 88.68  & 0.785\\
\hline
\parbox[t]{2mm}{\multirow{3}{*}{\rotatebox[origin=c]{90}{Easy}}}& \BigGANtrunc &  Library &      17.4\% &      22.0\% &     12.8\% &  2.1 &  13.3--21.6 & 0.049 & 98.45&0.695 \\
& \BigGANnotrunc & Library &      22.9\% &      28.1\% &     17.6\% &  2.1 &  18.9--27.2 & 0.029 & 78.49& 0.814\\
& SN-GAN  &   Library &      13.6\% &      15.1\% &     12.1\% &  1.9 &  10.0--17.5 & 0.043 & 94.89&0.814\\
\hline
\hline
\parbox[t]{2mm}{\multirow{3}{*}{\rotatebox[origin=c]{90}{Hard}}} & \BigGANtrunc &  French Horn &       7.3\% &       9.0\% &      5.5\% &  1.8 &   4.0--11.2 & 0.031 & 78.21&0.732\\
& \BigGANnotrunc & French Horn &       6.9\% &       8.6\% &      5.2\% &  1.4 &    4.3--9.9 & 0.042 & 96.18&0.757\\
& SN-GAN  &   French Horn &       3.6\% &       5.0\% &      2.2\% &  1.0 &    1.8--5.9 & 0.156 & 196.12&0.674\\
\hline
\parbox[t]{2mm}{\multirow{3}{*}{\rotatebox[origin=c]{90}{Hard}}}& \BigGANtrunc & Baseball Player &       1.9\% &       1.9\% &      1.9\% &  0.7 &    0.8--3.5 & 0.049 & 91.31&0.853\\
& \BigGANnotrunc &  Baseball Player &       2.2\% &       3.3\% &      1.2\% &  0.6 &    1.3--3.5 & 0.026 & 76.71&0.838\\
& SN-GAN  & Baseball Player &       2.8\% &       3.6\% &      1.9\% &  1.5 &    0.8--6.2 & 0.052 & 105.82&0.785\\
\hline
\end{tabular}}
\caption{\systeminf on three models trained on ImageNet and conditionally sampled on five classes. BigGAN routinely outperforms SN-GAN. \BigGANtrunc and \BigGANnotrunc are not separable.}
\label{table:imagenet-5}
\end{table}

\begin{table}[h]
\centering
\resizebox{\textwidth}{!}{\begin{tabular}[tb]{clccccccc}
GAN & \textbf{\systeminf (\%)} & Fakes Error & Reals Error & Std. & 95\% CI & KID & FID & Precision\\
\hline
\trunc &        23.3\% &      28.2\% &     18.5\% &  1.6 &  20.1--26.4 & 0.005 & 62.9 & 0.982\\
PROGAN &        14.8\% &      18.5\% &     11.0\% &  1.6 &  11.9--18.0 & 0.001 & 53.2 &  0.990\\
BEGAN &        14.5\% &      14.6\% &     14.5\% &  1.7 &  11.3--18.1 & 0.056 & 96.2 & 0.326\\
WGAN-GP &        13.2\% &      15.3\% &     11.1\% &  2.3 &   9.1--18.1 & 0.046 & 104.0 & 0.654\\
\hline
\end{tabular}}
\caption{Four models on CIFAR-10. \trunc can generate realistic images from \taskcifar.}
\label{table:cifar}
\end{table}

\subsubruninhead{\taskcifar}

For the difficult task of unconditional generation on \taskcifar, we use the same four model architectures in Experiment 1: \taskceleba. Table \ref{table:cifar} shows that \systeminf was able to separate \trunc from the earlier BEGAN, WGAN-GP, and ProGAN, indicating that StyleGAN is the first among them to make human-perceptible progress on unconditional object generation with \taskcifar. 

\textbf{Comparison to automated metrics}. Spearman rank-order correlation coefficients on all four GANs show medium, yet statistically insignificant, correlations with KID ($\rho=-0.600, p=0.40$) and FID ($\rho=0.600, p=0.40$) and precision ($\rho=-.800, p=0.20$).


\subsection{Related work}

\runinhead{Cognitive psychology.} We leverage decades of cognitive psychology to motivate how we use stimulus timing to gauge the perceptual realism of generated images. It takes an average of $150$ms of focused visual attention for people to process and interpret an image, but only $120$ms to respond to faces because our inferotemporal cortex has dedicated neural resources for face detection~\cite{rayner2009eye,chellappa2010face}. Perceptual masks are placed between a person's response to a stimulus and their perception of it to eliminate post-processing of the stimuli after the desired time exposure~\cite{sperling1963model}. Prior work in determining human perceptual thresholds~\cite{greene2009briefest} generates masks from their test images using the texture-synthesis algorithm~\cite{portilla2000parametric}. We leverage this literature to establish feasible lower bounds on the exposure time of images, the time between images, and the use of noise masks.

\runinhead{Success of automatic metrics.} Common generative modeling tasks include realistic image generation~\cite{goodfellow2014generative}, machine translation~\cite{bahdanau2014neural}, image captioning~\cite{vinyals2015show}, and abstract summarization~\cite{mani1999advances}, among others. These tasks often resort to automatic metrics like the Inception Score (IS)~\cite{salimans2016improved} and Fr\'echet Inception Distance (FID)~\cite{heusel2017gans} to evaluate images and BLEU~\cite{papineni2002bleu}, CIDEr~\cite{vedantam2015cider} and METEOR~\cite{banerjee2005meteor} scores to evaluate text. While we focus on how realistic generated content appears, other automatic metrics also measure diversity of output, overfitting, entanglement, training stability, and computational and sample efficiency of the model~\cite{borji2018pros,lucic2018gans,barratt2018note}. Our metric may also capture one aspect of output diversity, insofar as human evaluators can detect similarities or patterns across images. Our evaluation is not meant to replace existing methods but to complement them.

\runinhead{Limitations of automatic metrics.} Prior work has asserted that there exists coarse correlation of human judgment to FID~\cite{heusel2017gans} and IS~\cite{salimans2016improved}, leading to their widespread adoption. Both metrics depend on the Inception-v3 Network~\cite{szegedy2016rethinking}, a pretrained ImageNet model, to calculate statistics on the generated output (for IS) and on the real and generated distributions (for FID). The validity of these metrics when applied to other datasets has been repeatedly called into question~\cite{barratt2018note, rosca2017variational, borji2018pros, ravuri2018learning}. Perturbations imperceptible to humans alter their values, similar to the behavior of adversarial examples~\cite{kurakin2016adversarial}. Finally, similar to our metric, FID depends on a set of real examples and a set of generated examples to compute high-level differences between the distributions, and there is inherent variance to the metric depending on the number of images and which images were chosen---in fact, there exists a correlation between accuracy and budget (cost of computation) in improving FID scores, because spending a longer time and thus higher cost on compute will yield better FID scores~\cite{lucic2018gans}. Nevertheless, this cost is still lower than paid human annotators per image.

\runinhead{Human evaluations.} Many human-based evaluations have been attempted to varying degrees of success in prior work, either to evaluate models directly~\cite{denton2015deep,olsson2018skill} or to motivate using automated metrics~\cite{salimans2016improved,heusel2017gans}. Prior work also used people to evaluate GAN outputs on CIFAR-10 and MNIST and even provided immediate feedback after every judgment~\cite{salimans2016improved}. 
They found that generated MNIST samples have saturated human performance --- i.e.~people cannot distinguish generated numbers from real MNIST numbers, while still finding $21.3\%$ error rate on CIFAR-10 with the same model~\cite{salimans2016improved}. This suggests that different datasets will have different levels of complexity for crossing realistic or hyper-realistic thresholds. The closest recent work to ours compares models using a tournament of discriminators~\cite{olsson2018skill}. Nevertheless, this comparison was not yet rigorously evaluated on humans nor were human discriminators presented experimentally. The framework we present would enable such a tournament evaluation to be performed reliably and easily.


\subsection{Discussion}
\label{sec:discussion}

\runinhead{Envisioned Use.} 
We created \system as a turnkey solution for human evaluation of generative models. Researchers can upload their model, receive a score, and compare progress via our online deployment. 
During periods of high usage, such as competitions, a retainer model~\cite{bernstein2011crowds} enables evaluation using \systeminf in $10$ minutes, instead of the default $30$ minutes.

\runinhead{Limitations.}
Extensions of \system may require different task designs.
In the case of text generation (translation, caption generation), \systemstair will require much longer and much higher range adjustments to the perceptual time thresholds~\cite{krishna2017dense,weld2015artificial}. In addition to measuring realism, other metrics like diversity, overfitting, entanglement, training stability, and computational and sample efficiency are additional benchmarks that can be incorporated but are outside the scope of this paper. Some may be better suited to a fully automated evaluation~\cite{borji2018pros,lucic2018gans}. Similar to related work in evaluating text generation~\cite{hashimoto2019unifying}, we suggest that diversity can be incorporated using the automated recall score measures diversity independently from precision $F_{1/8}$~\cite{sajjadi2018assessing}.

\runinhead{Conclusion.}
\system provides two human evaluation benchmarks for generative models that (1)~are \textbf{grounded} in psychophysics, (2)~provide task designs that produce \textbf{reliable} results, (3)~\textbf{separate} model performance, (4)~are cost and time \textbf{efficient}. We introduce two benchmarks: \systemstair, which uses time perceptual thresholds, and \systeminf, which reports the error rate sans time constraints. We demonstrate the efficacy of our approach on image generation across six models \{StyleGAN, SN-GAN, BigGAN, ProGAN, BEGAN, WGAN-GP\}, four image datasets \{\taskceleba, \taskffhq, \taskcifar, \taskimagenet\}, and two types of sampling methods \{with, without the truncation trick\}. 

\section{Conclusion}
\label{sec:5}
Popular culture has long depicted vision as a primary interaction modality between people and machines; vision is a necessary sensing capability for humanoid robots such as \textit{C-3PO} from ``Star Wars'', \textit{Wall-E} from the eponymous film, and even disembodied Artificial Intelligence such as \textit{Samantha} the smart operating system from the movie ``Her''. 
These fictional machines paint a potential real future where machines can tap into the expressive range of non-intrusive information that Computer Vision affords.
Our expressions, gestures, and relative position to objects carry a wealth of information that intelligent interactive machines can use, enabling new applications in domains such as healthcare~\cite{haque2020illuminating}, sustainability~\cite{jean2016combining}, human-interpretable actions~\cite{dragan2013legibility}, and mixed-initiative interactions~\cite{horvitz1999principles}.

While Human-Computer Interaction (HCI) researchers have long discussed and debated what human-AI interaction should look like~\cite{10.1145/267505.267514,horvitz1999principles}, we have rarely provided concrete, immediately operational goals to Computer Vision researchers. Instead, we've largely left this job up to the vision community itself, which has produced a variety of immediately operational tasks to work on. These tasks play an important role in the AI community; some of them ultimately channel the efforts of thousands of AI researchers and set the direction of progress for years to come. The tasks range from object recognition~\cite{deng2009imagenet}, to scene understanding~\cite{visualgenome}, to explainable AI~\cite{adadi2018peeking}, to interactive robot training~\cite{thomaz2008teachable}. And while many such tasks have been worthwhile endeavors, we often find that the models they produce don't work in practice or don't fit end-users' needs as hoped~\cite{mitchell2019model,buolamwini2018gender}.

If the tasks that guide the work of thousands of AI researchers do not reflect the HCI community's understanding of how humans can best interact with AI-powered systems, then the resulting AI-powered systems will not reflect it either. 
We therefore believe there is an important opportunity for HCI and Computer Vision researchers to begin closing this gap by collaborating to directly integrate HCI's insights and goals into immediately actionable vision tasks, model designs, data collection protocols, and evaluation schemes. One such example of this type of work is the HYPE benchmark mentioned earlier in this chapter~\cite{zhou2019hype}, which aimed to push GAN researchers to focus directly on a high-quality measurement of human perception when creating their models. Another is the approach taken by the social strategies project mentioned earlier in this chapter~\cite{park2019ai}, which aimed to push data collection protocols to consider social interventions designed to elicit volunteer contributions.

What kind of tasks might HCI researchers work to create? 
First, explainable AI aims to help people understand how computer vision models work, but methods are developed without real consideration of how humans will ultimately use explanations to interact with them. HCI researchers might propose design choices for how to introduce and explain vision models grounded in human subjects experiments~\cite{khadpe2020conceptual,buccinca2020proxy}. 
Second, perceptual robotics can learn to complete new tasks by incorporating human rewards, but do not consider how people actually want to provide feedback to robots~\cite{thomaz2008teachable}. If we want robots to incorporate an understanding of how humans want to give feedback when deployed, then HCI researchers might propose new training paradigms with ecologically valid human interactions. 
Third, multi-agent vision systems~\cite{jain2020cordial} are developed that ignore key aspects of human psyche, such as choosing to perform non-optimal behaviors, despite foundational work in HCI noting the perils of such assumptions in AI planning~\cite{suchman1987plans}. Without incorporating human behavioral priors, these multi-agents systems work well when collaborating between AI agents but fail when one of the agents is replaced by a human~\cite{carroll2019utility}. If we want multi-agent vision systems that understand biases that people have when performing actions, then HCI researchers might propose human-AI collaboration tasks and benchmarks in which agents are forced to encounter realistic human actors (indeed, non-vision work has begun to move in this direction~\cite{kwon2020humans}).

\runinhead{Acknowledgement.} 
The first project was supported by the National Science Foundation award 1351131.
The second project was partially funded by the Brown Institute of Media Innovation and by Toyota Research Institute (``TRI''). 
The third project was partially funded by a Junglee Corporation Stanford Graduate Fellowship, an Alfred P. Sloan fellowship and by TRI.
This chapter solely reflects the opinions and conclusions of its authors and not TRI or any other Toyota entity.

\bibliographystyle{plain}
\bibliography{references}

\end{document}